\documentclass[10pt,twocolumn,letterpaper]{article}

\usepackage{iccv}
\usepackage{times}
\usepackage{epsfig}
\usepackage{graphicx}
\usepackage{amsmath}
\usepackage{amssymb}
\usepackage{multirow}
\usepackage{comment}
\usepackage{mathrsfs}

\usepackage{color}
\usepackage{subfigure}
\usepackage{multirow}
\usepackage{multicol}

\newcommand{\ruleh}{\unskip\ \hrule\ }

\usepackage[breaklinks=true,bookmarks=false]{hyperref}

\iccvfinalcopy 

\def\iccvPaperID{****} 

\begin{document}

\title{Handwritten Chinese Font Generation with Collaborative Stroke Refinement}

\author{Chuan Wen$^1$ \quad Jie Chang$^1$ \quad  Ya Zhang$^1$ \quad Siheng Chen$^2$ \quad Yanfeng Wang$^1$ \quad Mei Han$^3$ \quad Qi Tian$^4$\\
$^1$Cooperative Madianet Innovation Center, Shanghai Jiao Tong University \\ $^2$Mitsubishi Electric Research Laboratories \\ $^3$PingAn Technology US Research Lab \\ $^4$Huawei Technologies Noah's Ark Lab\\
{\tt\small \{alvinwen, j\_chang, ya\_zhang, wangyanfeng\}@sjtu.edu.cn \quad sihengc@andrew.cmu.edu} \\ {\tt\small hanmei613@pingan.com.cn \quad tian.qi1@huawei.com}
}

\maketitle

\begin{abstract}
   Automatic character generation is an appealing solution for new typeface design, especially for Chinese typefaces including over 3700 most commonly-used characters. This task has two main pain points: (i) handwritten characters are usually associated with thin strokes of few information and complex structure which are error prone during deformation;
    (ii) thousands of characters with various shapes are needed to synthesize based on a few manually designed characters. To solve those issues, we propose a novel convolutional-neural-network-based model with three main techniques: \emph{collaborative stroke refinement}, using collaborative training strategy to  recover the missing or broken strokes; \emph{online zoom-augmentation}, taking the advantage of the content-reuse phenomenon to reduce the size of training set; and \emph{adaptive pre-deformation}, standardizing and aligning the characters. The proposed model needs only $750$ paired training samples; no pre-trained network, extra dataset resource or labels is needed. Experimental results show that the proposed method significantly outperforms the state-of-the-art methods under the practical restriction on handwritten font synthesis. 
\end{abstract}

\begin{figure}[htb]
\centering
\includegraphics[width=0.45\textwidth]{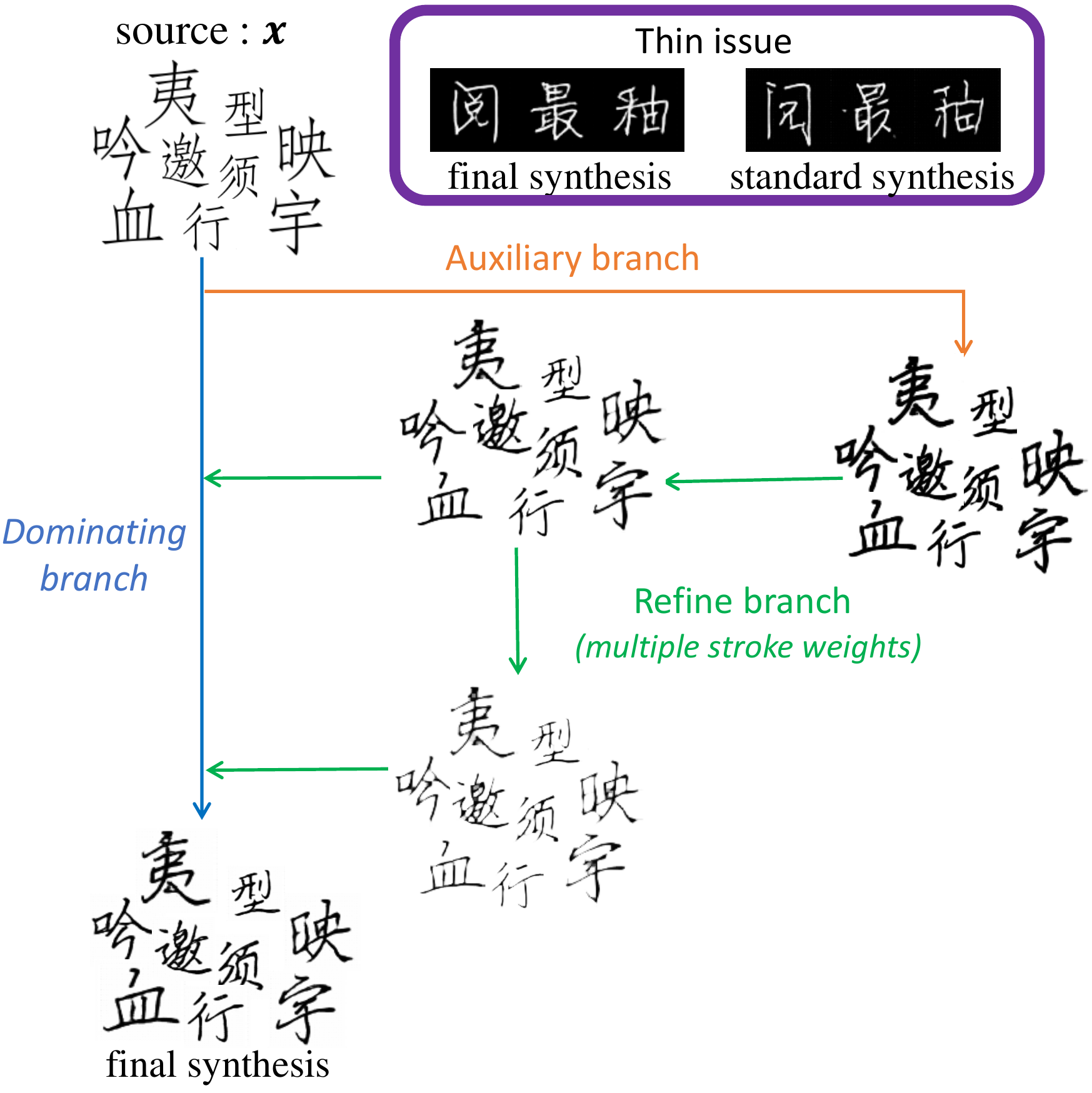}
\caption{Collaborative stroke refinement handles the thin issue. Handwritten characters usually have various stroke weights. Thin strokes have lower fault tolerance in synthesis, leading to distortions. Collaborate stroke refinement adopts collaborative training. An auxiliary branch is introduced to capture various stroke weights, guiding the dominating branch to solve the thin issue.}
\label{fig:morpological-motivation}
\end{figure}

\begin{figure}[h]
\centering
\includegraphics[width=0.45\textwidth]{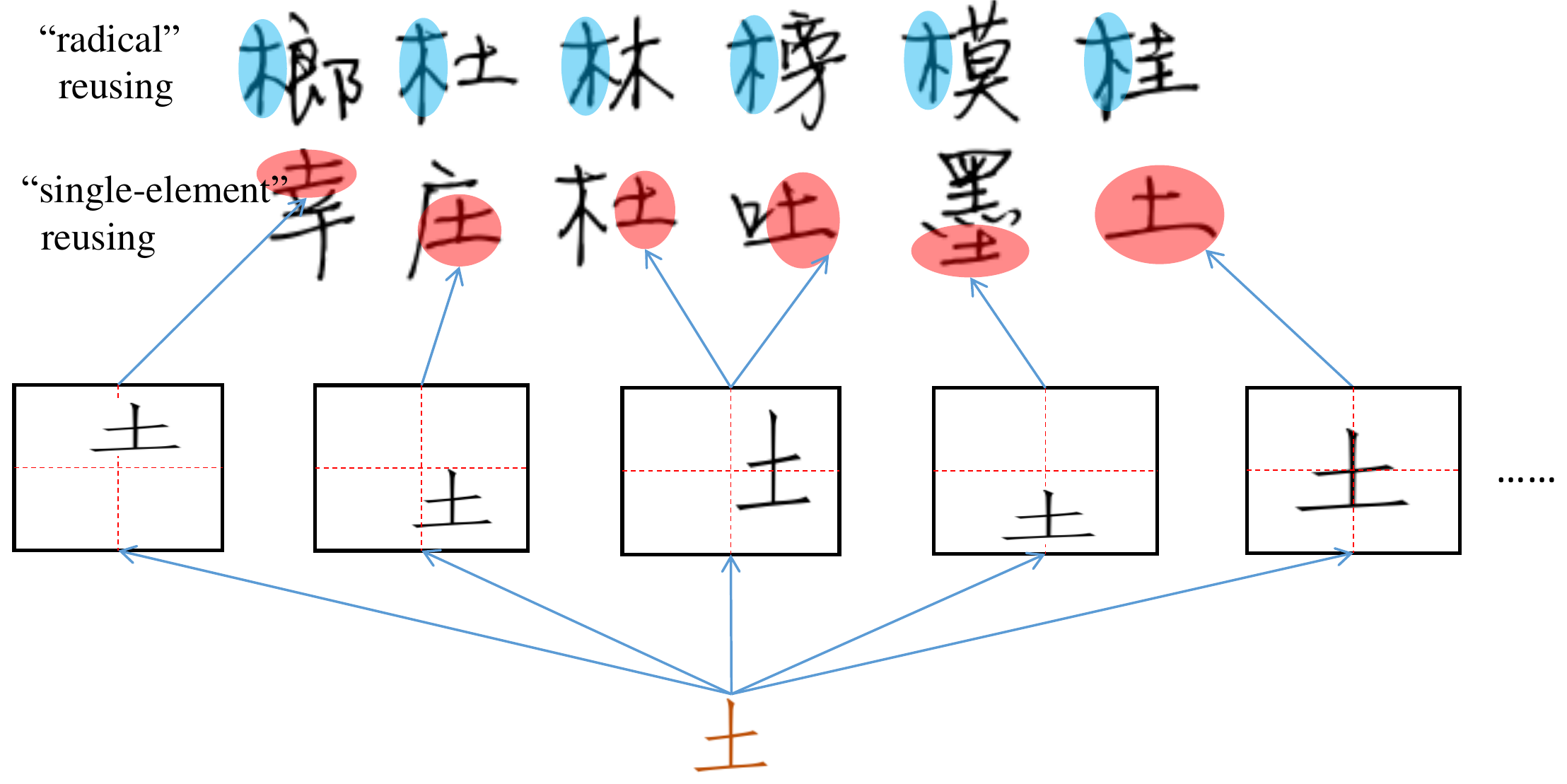}
\caption{Online zoom-augmentation exploits the content-reuse phenomenon in Chinese characters. A same radical can be reused in various characters at various locations. Online zoom-augmentation implicitly model the variety of locations and sizes from elementary characters without decomposition, significantly reducing the number of training samples.}
\label{fig: intro-2}
\end{figure}
\section{Introduction}
Chinese language consists of more than 8000 characters, among which about 3700 are frequently used. Designing a new Chinese font involves considerable tedious manual efforts which are often prohibitive. A desirable solution is to automatically complete the rest of vocabulary given an initial set of manually designed samples, i.e. \emph{Chinese font synthesis}. Inspired by the recent progress of neural style transfer~\cite{chang2018pairedcyclegan, chen2018cartoongan, gatys, Sanakoyeu_2018_ECCV, wang2018learning, CycleGAN}, several attempts have been recently made to model font synthesis as an image-to-image translation problem~\cite{changchinese,sun2017learning,zhang2018separating}. Unlike neural style transfer, font transfer is a low fault-tolerant task because any misplacement of strokes may change the semantics of the characters ~\cite{changchinese}. 
So far, two main streams of approaches have been explored. One addresses the problem with a bottleneck CNN structure~\cite{Rewrite,zi2zi,changchinese,lyu2017auto,jiang2017dcfont} and the other proposes to disentangle representation of characters into \emph{content} and \emph{style}~\cite{sun2017learning,zhang2018separating}. 
While promising results have been shown in font synthesis,  most existing methods require an impracticable large-size initial set to train a model.
For example, ~\cite{Rewrite,zi2zi,changchinese,lyu2017auto} require 3000 paired characters for training supervision, which requires huge labor resources. 
Several recent methods~\cite{jiang2017dcfont, sun2017learning, zhang2018separating} explore learning with less paired characters; however, these methods heavily depend on extra labels or other pre-trained networks.
Table~\ref{tab:comp_methods} provides a comparative summary of the above methods. 

\begin{table*}[!t]\footnotesize{}
    \centering
    \caption{Comparison of {\em Our} with existing Chinese font transfer methods based on CNN.}
    \begin{tabular}{c|c|c|c}
    \hline
        Methods & Requirements of data resources & Dependence of extra label? & Dependence of pre-trained network?\\
    \hline Rewrite~\cite{Rewrite} & \multirow{3}{3cm}{3000 paired samples.} & \multirow{3}{1cm}{None} & \multirow{4}{3cm}{None}\\
    \cline{1-1} AEGN~\cite{lyu2017auto} & & &  \\
    \cline{1-1} HAN~\cite{changchinese} & & & \\
    \cline{1-3} zi2zi~\cite{zi2zi} & Hundreds of fonts (each containing 3000 samples). & \multirow{2}{1.5cm}{Class-label for each font.} & \\
    \cline{1-2}\cline{4-4} DCFont~\cite{jiang2017dcfont} & Hundreds of fonts (each containing 755 samples). & & Pre-trained font-classifier based on Vgg16\\
    \cline{1-4} SA-VAE~\cite{sun2017learning} & \multirow{2}{4cm}{Hundreds of fonts (each containing 3000 samples).} & 133-bit structural code for each character. & Pre-trained character recognition network\\
    \cline{1-1}\cline{3-4} EMD~\cite{zhang2018separating} & &
14
 \multirow{2}{1cm}{None} & \multirow{2}{3cm}{None}\\
    \cline{1-2} \textbf{Ours} & Only 755 paired samples & & \\
    \hline
    \end{tabular}%
    \vspace{-10pt}
  \label{tab:comp_methods}%
\end{table*}%

Most of the above methods focus on printed font synthesis. Compared with printed typeface synthesis, \emph{handwritten} font synthesis is a much more challenging task. First, its strokes are thinner especially in the joints of strokes. Handwriting fonts are also associated with irregular structures and are hard to be modeled. So far there is no satisfying solution for \emph{handwritten} font synthesis. 

This paper aims to synthesize  more realistic handwritten fonts with fewer training samples. There are two main challenges: (i) handwritten characters are usually associated with thin strokes of few information and complex structure which are error prone during deformation; (see Figure~\ref{fig:morpological-motivation}) and (ii) thousands of characters with various shapes are needed to synthesize based on a few manually designed characters (see Figure~\ref{fig: intro-2}). To solve the first issue, we propose \textit{collaborative stroke refinement} that  using collaborative training strategy to recover the missing or broken strokes. Collaborative stroke refinement uses an
auxiliary branch to generate the syntheses with various stroke weights, guiding the dominating branch to capture characters with various stroke weights. To solve the second issue, we fully exploit the content-reuse phenomenon in Chinese characters; that is, the same radicals may present in various characters at various locations. Based on this observation, we propose online zoom-augmentation, which 
synthesizes a complicated character by a series of elementary components. The proposed networks can focus on
learning varieties of locations and ratios of those elementary components. This significantly reduces the size of training samples. To make the proposed networks easy to capture the deformation from source to target, we further propose adaptive pre-deformation, which learns the size and scale deformation to standardize characters. This allows the proposed networks focus on learning high-level style deformation.

\begin{figure*}[htb]
\centering
\includegraphics[width=0.95\textwidth, height=6cm]{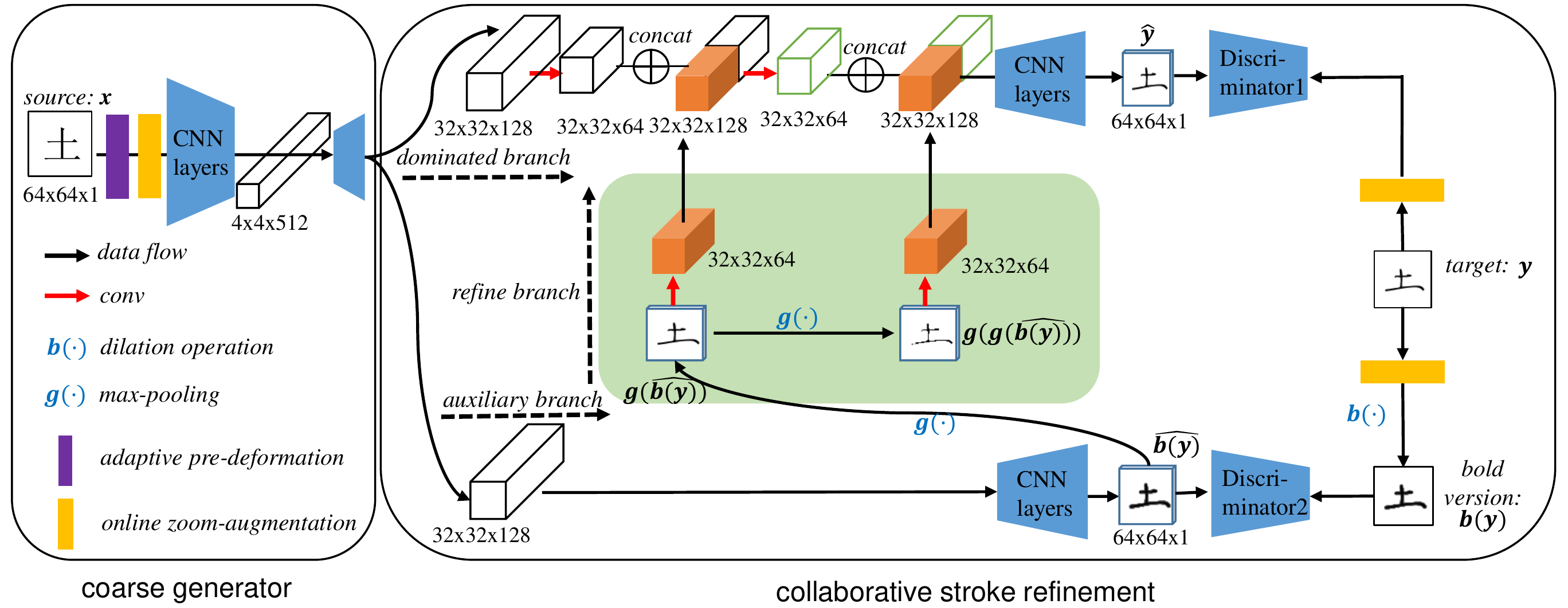}
\caption{Schematic of proposed model including \textit{coarse generator} and \textit{collaborative stroke refinement}. In coarse generator and \textit{source} image $x$ is transformed to $32\times32\times128$ feature map after adaptive pre-deformation, online zoom-augmentation $\mathcal{A}$. In \textit{collaborative stroke refinement}, \textit{auxiliary branch} generates $\hat{b(y)}$, which uses $b(y)$ as ground-truth. Simultaneously, \textit{dominating branch} generates $\hat{y}$ guided by our desired target $y$.  And in refinement branch, $\hat{b(y)}$ is eroded to a thin version $g(\hat{b(y)})$ and feature maps extracted from it flow to \textit{dominating branch} as compensation. Besides, skip-connections are used between \textit{coarse generator} and the \textit{refine branch}.}
\label{fig: framework}
\end{figure*}

Combining the above methods, we proposed an end-to-end model (as shown in Fig.\ref{fig: framework}), generating high-quality characters with only 750 training samples. No pre-trained network, or labels is needed. We verify the performance of the proposed model on several Chinese fonts including both \emph{handwritten} and \emph{printed} fonts. The results demonstrate the significant superiority of our proposed method over the state-of-the-art Chinese font generation models.

The main contributions are summarized as follows.
\begin{itemize}
    \item We propose an end-to-end model to synthesize handwritten  Chinese  font  given  only  750 training samples;
    \item We propose collaborative stroke refinement, handling the thin issue; online zoom-augmentation, enabling learning with fewer training samples; and adaptive pre-deformation, standardizing  and aligning the characters;
    \item Experiments show that the proposed networks are far superior to baselines in respect of visual effects, RMSE metric and user study.
\end{itemize}


\section{Related Work}
Most image-to-image translation tasks such as art-style transfer~\cite{chen2017stylebank,johnson2016perceptual}, coloring~\cite{cao2017unsupervised,zhang2016colorful}, super-resolution~\cite{ledig2017photo}, dehazing~\cite{ancuti2018ntire} have achieved the progressive improvement since the raise of CNNs.  Recently, several works model font glyph synthesis as a supervised image-to-image translation task mapping the character from one typeface to another by CNN-framework analogous to an encoder/decoder backbone~\cite{Rewrite,zi2zi,Azadi_2018_CVPR,changchinese,jiang2017dcfont,lyu2017auto}. Specially, generative adversarial networks (GANs)~\cite{goodfellow2014generative,mirza2014conditional,radford2015unsupervised} are widely adopted by them for obtaining promising results. Differently, EMD~\cite{zhang2018separating} and SA-VAE~\cite{sun2017learning} are proposed to disentangle the style/content representation of Chinese fonts for a more flexible transfer.

Preserving the content consistency is important in font synthesis task. ``From A to Z'' assigned each Latin alphabet a one-hot label into the network to constrain the content~\cite{upchurch2016z}. Similarly, SA-VAE~\cite{sun2017learning} embedded a pre-trained Chinese character recognition network~\cite{xiao2017building,zhong2016handwritten} into the framework to provide a correct content label. SA-VAE also embedded 133-bits coding denoting the structure/content-reuse knowledge of inputs, which must rely on the extra labeling related to structure and content beforehand. In order to synthesize multiple font, zi2zi~\cite{zi2zi} utilized a one-hot category embedding. Similarly, DCFont~\cite{jiang2017dcfont} involved a pre-trained 100-class font-classifier in the framework to provide a better style representation.
And recently, MC-GAN~\cite{Azadi_2018_CVPR} can synthesize
ornamented glyphs from a few examples. However, the generation of Chinese characters are very different from English alphabet no matter in complexity or considerations.
EMD~\cite{zhang2018separating} is the first model which can achieve good performance on new Chinese font transfer with a few samples, but it does  work poorly on handwritten fonts.

\section{Methodology}
The task of handwritten Chinese font generation is that given a source character, we aim to generate the same character in the target style. Mathematically, let $\mathcal{C} =  \{c_i\}_{i=0}^m$ be a image set of $m$ standardized Chinese characters, where each element $c_i$ represents a single Chinese character. Let the source set $\mathcal{X} =  \{ x_i = d_{\rm source}(c_i) | c_i \in  \mathcal{C} \}_{i}$ and the target set $\mathcal{Y} = \{ y_i = d_{\rm target}(c_i) | c_i \in  \mathcal{C} \}_{i}$ be two training image sets representing the characters $\mathcal{C}$ in two styles, where $d_{\rm source}(\cdot)$ denotes the deformation function in the source style and $d_{\rm target}(\cdot)$ denotes the deformation function in the target style. In the training phase, we train with the source-target pairs, $\mathcal{X}$ and $\mathcal{Y}$. In the testing phase, given an input image, $x = d_{\rm source}(c)$, we should produce a synthesis image, $y = d_{\rm target}(c)$. Since that we are blind to both deformation functions, the key of this task is to learn a mapping from $d_{\rm source}(\cdot)$ to $d_{\rm target}(\cdot)$ based on training sets $\mathcal{X}$ and $\mathcal{Y}$. To make it practical, we want the size of  the training set $m$ as small as possible.
Here we use deep neural networks to learn the mapping from $d_{\rm source}(\cdot)$ to $d_{\rm target}(\cdot)$

The proposed networks are illustrated in Fig. \ref{fig: framework}. The proposed networks consist of a \textit{coarse generator}, which is a shared encoder-decoder module to generate low resolution feature maps, \textit{collaborative stroke refinement}, which 
follows an encoder-decoder framework and generates the results. We further use adaptive pre-deformation and online zooming to augment the training data, which carefully exploit the specific properties of Chinese letters, enabling the proposed network to train with fewer training samples.


\subsection{Collaborative Stroke Refinement}
\label{sec: collaborative-stroke-refinement}
As illustrated in Fig.\ref{fig:morpological-motivation}, we see that generating bold strokes of handwritten fonts are usually easier than generating thin strokes of handwritten fonts, because thin strokes have fewer information and have lower fault tolerance in the reconstruction process. In practice, a single handwritten font may involve various stroke weights, making the synthesis challenging; see thin issue in Fig. \ref{fig:morpological-motivation}.

To solve this issue, we propose collaborative stroke refinement, which uses multitask learning and collaborative training strategy. Collaborative stroke refinement includes three branches: \textit{dominating branch}, \textit{auxiliary branch} and \textit{refine branch}; see Fig. \ref{fig: framework}.  The auxiliary branch learns stroke deformations between the source $x$ and a bolded target $b(y)$, where $b(\cdot)$ is a bolding function; the refine branch merges information from the auxiliary branch to the dominating branch; and finally, the dominating branch learns the stroke deformation between the source $x$ and the original target $y$ with auxiliary information from the refine branch. We train three branches simultaneously. 

\begin{figure}[htb]
\centering
\includegraphics{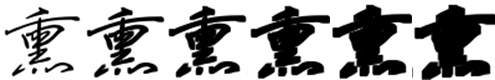}
\caption{Dilation may cause the strokes overlapping issue.}
\label{fig: erode_impact}
\end{figure}

\textbf{\textit{Auxiliary branch}.} The auxiliary branch  synthesizes $\hat{b(y)}$ to approximate the bolded target $b(y)$ based on the source $x$. The bolded target is obtained from original target by applying the morphological dilation operation:
\begin{equation*}\label{eq: eq3}
b(y)  = y \oplus e = \{z|(\hat{e})_z \cap y \neq \varnothing \},
\end{equation*}
where $e$ is the structuring element, $\hat{*}$ denotes reflecting all elements of $*$ about the origin of this set, $(*)_z$ denotes translating the origin of $*$ to point $z$. The synthesis $\hat{b(y)}$ has bold strokes and are robust to potential distortion. Since handwritten fonts' strokes are usually highly connected, simply using dilation may cause overlapping of strokes; see Fig.~\ref{fig: erode_impact}. Instead of using $\hat{b(y)}$ as the final synthesis, we use it as auxiliary information and output to the refine branch.

\textbf{\textit{Refine branch}.} Refine branch merges information from the auxiliary branch to the dominating branch. To allow the dominating branch to capture various stroke weights, we aim to make $\hat{b(y)}$ thinner; however, directly using the erosion operation, which is the inverse of dilation, blocks the end-to-end training due to  erosion is not differentiable. To overcome this, we use max-pooling to mimic erosion, which is differentiable. We use max-pooling twice to produce two stroke-weight syntheses, $g(\hat{b(y)})$ and $g(g(\hat{b(y)}))$, where $g(\cdot)$ denote the max-pooling with stride $2$.  We input each of two syntheses, $g(\hat{b(y)})$ and $g(g(\hat{b(y)}))$, to the convolution layers separately and obtain feature maps that reflect the same style with various stroke weights.

\textbf{\textit{Dominating branch}.} 
The dominating branch collects auxiliary information from the refine branch and produces the final synthesis. The network structure is based on HAN with a generator and a hierarchical discriminator \cite{changchinese} (a discriminator to distinguish not only between the generated image and ground truth, but also between the feature maps, obtained from the first few layers before the output layer, and the feature maps of ground truth by conducting the same convolution), making the model converge faster and the results more smooth.  The input of the generator is the output feature maps of the coarse generator. We process it with a convolution layer and then concatenate it with $g(\hat{b(y)})$, which is the auxiliary information from the refine branch.  We next process the concatenation with another convolution layer and concatenate it with $g(g(\hat{b(y)}))$. Twice concatenations allow the dominating branch to be aware of various stroke weights. We put the concatenation to the CNN to reconstruct the final synthesis $\hat{y}$.  The hierarchical discriminator tries to distinguish the synthesis $\hat{y}$ and its feather maps from the ground truth $y$ and $y$'s feature maps, promoting the generator to converge faster.  The dominating branch fuses auxiliary information from the refine branch twice to capture various stroke weights, making the final synthesis robust to distortion and missing strokes.

The above three branches are trained simultaneously. The auxiliary branch generates $\hat{b(y)}$ to approximate the skeleton of the bolded target; the refine branch extracts features from various stroke-weight syntheses and pass them to the dominating branch; the dominating branch finally combines the auxiliary information from the refine branch and produces the final syntheses. This three-branch design follows similar favor of multitask learning and collaborative training strategy. It pushes the networks to learn the deformation from one style to another with various stroke weights and handles the thin-stroke issue. Furthermore, the idea of replacing the erosion operation by max-pooling makes the gradient flow through all the three branches; we thus call the method~\emph{collaborative stroke refinement}.


\subsection{Online Zoom-Augmentation}
\label{sec: online-zoom-augmentation}
Chinese characters have the content-reuse phenomenon; that is, the same radicals may present in various characters at various locations; see Fig. \ref{fig: intro-2}). In other words, a Chinese character can be decomposed into a series of elementary characters. Mathematically, the standardized character $c$ can be modeled as 
\begin{equation}
\label{eq:decompose}
\vspace{-4mm}
    c = \sum_{\ell} d_{\ell} (b_{\ell}),
\end{equation}
where $b_{\ell}$ is an elementary characters and $d_{\ell}(\cdot)$ is the deformation function associated with each elementary character. 

The shape of an arbitrary character $c$ could have huge variations and it is thus hard to learn deformation directly based on $c$; on the other hand, the shapes of the elementary character $b_{\ell}$ are limited and it is thus much easier to learn deformation directly based on $b_{\ell}$. The functionality of online zoom-augmentation is to explore this decomposition~\eqref{eq:decompose} adaptively. It has two main advantages: (i) it leverages the repetitiveness of radicals in Chinese characters and guides the networks to learn a few elementary structures of Chinese characters, leading to a significantly reduction of the training size; and (ii) it guides the networks learn characters at a variety of locations and sizes, making the networks robust. To our best knowledge, no CNN-based methods leverage this domain knowledge except SA-VAE~\cite{sun2017learning}. However, SA-VAE explicitly models this as a pre-labeling 133-bits code embedded into the model.  Instead,we carefully select $750$ training samples as the elementary character $b_{\ell}$; see details in Section~\ref{sec: sample-selection}

We further use various positions and scales to train the element-wise deformation operator $d_{\ell}(\cdot)$. Specifically, when a paired image ${x,y}$ is fed into the model, we zoom the centered character region to change the aspect ratio. We then translate the zoomed character region horizontally or vertically. Assuming the ratio of each character region is $h:w$, the zoomed result with $\frac{h}{2}:w$ will be vertically translated in the image (mostly translated to the upper/middle/bottom position), while the zoomed result with $h:\frac{w}{2}$ will be horizontally translated (mostly translated to the left/middle/right position). Additionally, the zoomed result with $\frac{h}{2}:\frac{w}{2}$ is translated to any possible position. Essentially, $d_{\ell}(\cdot)$ captures the mapping between $b_{\ell}$ and radicals of arbitrary character. Augmented training samples guide the networks to learn a variety of location patterns.

\subsection{Adaptive Pre-deformation}
\label{sec: Adaptive-Pre-deformation}
The overall task is to learn a mapping from $x = d_{\rm source}(c)$ to $y = d_{\rm target}(c)$ for arbitrary character $c$. In Section~\ref{sec: online-zoom-augmentation}, we decompose the character $c$. We can further decompose the deformations, $d_{\rm source}(\cdot)$ and $d_{\rm target}(\cdot)$, to ease the learning process. The main intuition is that some basic deformations, such as resize and rescale, are easy to learn. We can use a separate component to focus on learning those basic deformations and the main networks can then focus on learning complicated deformations. Mathematically, a general deformation function can be decomposed as
\begin{equation*}
    d(\cdot)  = d_{\rm complex }(  d_{\rm rescale }( d_{\rm reszie }(  \cdot  )    )   ).
\end{equation*}
The functionality of adaptive pre-deformation is to learn $d_{\rm rescale }( \cdot    )$ and $d_{\rm reszie }(  \cdot  )$, such that the main networks can focus on learning $d_{\rm complex }(\cdot)$.
 
All paired ${x^i, y^{i}}$ from $S_D$ construct the source image set $\mathcal{X}$ and the target image set $\mathcal{Y}$. We calculate the average character-region proportion $r_1$ and average character's height-width ration $r_2$ for $\mathcal{Y}$. First we find the minimum character bounding box $b_{i}$ of each $y^{i}$, the height and width of $b_i$ is respectively $h_i$ and $w_i$. So,
\begin{equation*}
\vspace{-2mm}
r_{1} = \frac{1}{N}\sum_{i}^{N}\frac{h_i\cdot w_i}{64\times64}, r_{2} = \frac{1}{N}\sum_{i}^{N}\frac{h_i}{w_i};
\label{eq: eq1}
\end{equation*}
where $N=750$. According to the above two statistics, we then pre-deformed each $x^i$ to align its character region with $y^i$. The deformed result $\hat{x^i}$ is:
\begin{equation*}
\vspace{-3mm}
\hat{x^i} = d_{\rm rescale}(d_{\rm resize}(x^i)),
\label{eq: eq2}
\end{equation*}
where $d_{\rm resize}(\cdot)$ and $d_{\rm rescale}(\cdot)$ denote the size-deformation and scale-deformation, respectively. The character skeleton of $x^{i}$ is then roughly aligned with $y^{i}$. Here by  pre-deformation, the skeleton of source character is roughly aligned with the corresponding target character, which reduces the transfer-difficulty. Specifically, the model does not fit the size and can focus on learning stroke distortions.

\vspace{-2mm}
\subsection{Losses}
We next describe loss functions optimized in our model. There are 4 loss terms divided into two groups, $L_{pixel}^{1}+L_{cGAN}^{1}$ and $L_{pixel}^2+L_{cGAN}^{2}$. 
\begin{equation*}
\label{eq: eq6}
\vspace{-3mm}
    L_{pixel}^{1}(G_1) =
    \mathbb{E}_{(x,y)}[-y\cdot(\log(\hat{y}))
 -(1-y)\cdot\log(1-\hat{y})],
\end{equation*}

\begin{eqnarray}
\vspace{-5mm}
\label{eq: eq7}
\begin{split}
    L_{cGAN}^{1}(G_1,D_1) &=& \mathbb{E}_{(x,y)}[\log D_{1}(x, y)]
\\ 
    && + \mathbb{E}[1-\log D_{1}(x, G_{1}(x)],
\end{split}
\end{eqnarray}
where $y$ is the character with target font, $(x,y)$ is pair-wise samples, $x\sim p_{source\_domain}(x)$ and $y\sim p_{target\_domain}(y)$. $D_1$ is the \textit{discriminator} 1. $G_1$ includes \textit{coarse generator} and the \textit{refine branch}.
\begin{equation*}
\vspace{-5mm}
\begin{split}
\label{eq: eq8}
    L_{pixel}^{2}(G_2) = 
    \mathbb{E}_{(x,b(y))}[-b(y)\cdot(\log(\hat{b(y)}))
    \\ -(1-b(y))\cdot\log(1-\hat{b(y)})],
\end{split}
\end{equation*}

\begin{eqnarray}
\vspace{-5mm}
\label{eq: eq9}
\begin{split}
    L_{cGAN}^{2}(G_2, D_2) &=& \mathbb{E}_{(x,b(y))}[\log D_{2}(x, b(y))]\\
    && +\mathbb{E}[1-\log D_{2}(x, G_{2}(x)],
\end{split}
\end{eqnarray}
where $b(y)$ is the \textbf{bold} version character of target font. $D_2$ is the \textit{discriminator} 2. $G_2$ only contains the \textit{refine branch}. The whole network is jointly optimized by $L_{pixel}^{1}+L_{cGAN}^{1}+L_{pixel}^2+L_{cGAN}^{2}$. It is notable that all $x$, $y$ and $b(y)$ here are augmented by $d_{\ell}(\cdot)$ and pre-deformation function $d_{\rm resize}(\cdot)$, $d_{\rm rescale}(\cdot)$. We just write this for simplicity.


\section{Experiments}
We conduct extensive experiments to validate the proposed method and compare it with two state-of-the-art methods, HAN\cite{changchinese} and EMD\cite{zhang2018separating}


\begin{figure}[htb]
\centering
\includegraphics[width=0.35\textwidth]{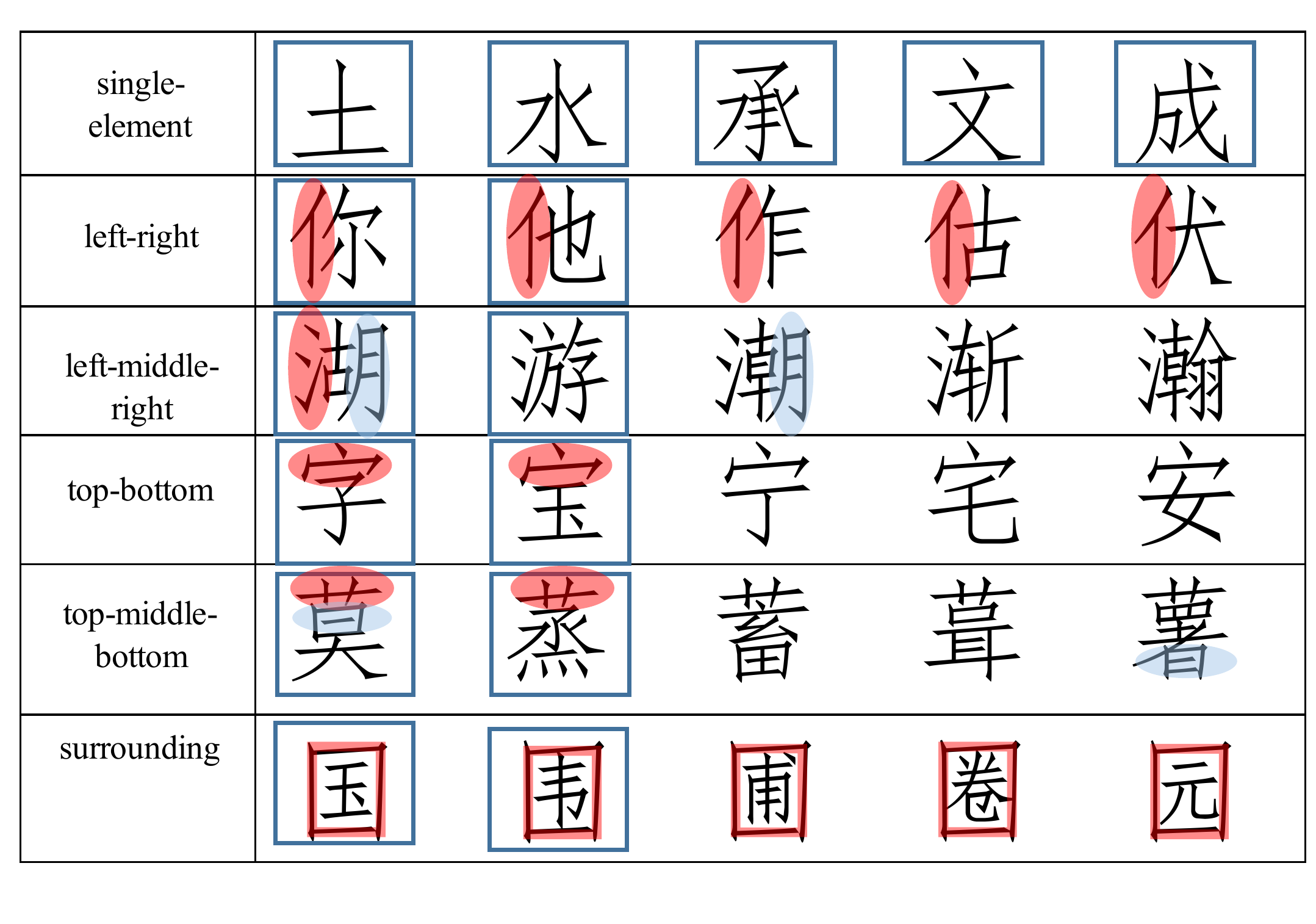}
\caption{Examples of ``compound'' and ``single-element'' characters. Some radicals are marked in red.}
\label{fig: selection}
\end{figure}

\begin{figure}[htb]
\centering
\includegraphics[width=0.4\textwidth]{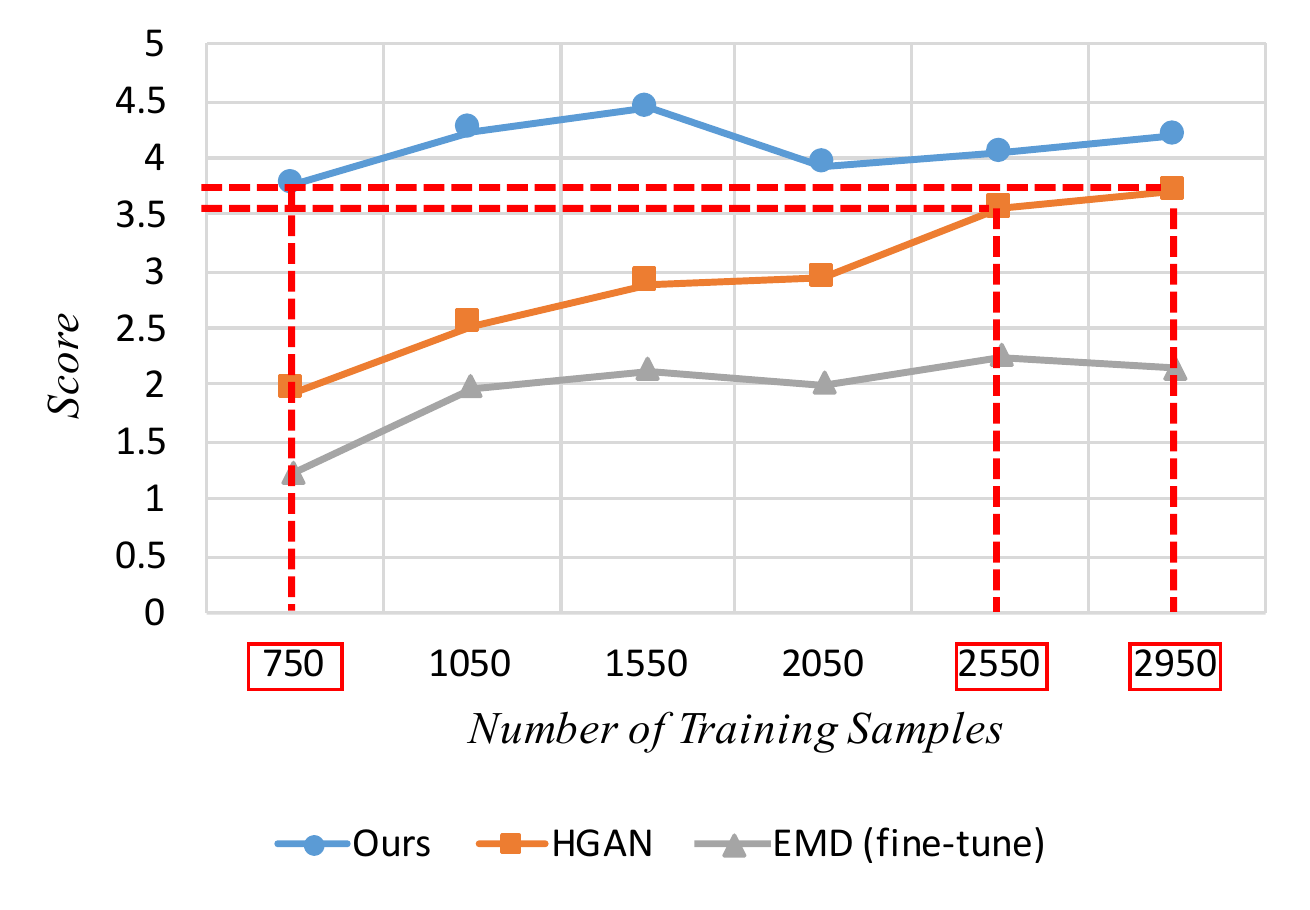}
\caption{In the user study, our method achieves the close subjective score with much less training set compared with baselines. }
\label{fig: table}
\end{figure}



\subsection{Training Sample Selection}
\label{sec: sample-selection}
For Chinese characters, ``compound'' characters appear as obvious layout structure (e.g. left-right, top-bottom, surrounding, left-middle-right or top-middle-bottom), while ``single-element'' characters cannot be structural decomposed. Both radicals and ``single-element'' characters are known as basic units to construct all characters (see Fig. \ref{fig: selection}). 
Many compound Chinese characters share the same basic units in themselves, which means though over 8000 characters in Chinese language, there are only rather limited basic units (including 150 radicals and about 450 ``single-element'' characters). Based on this prior knowledge, a small set of characters are selected as training samples. We select 450 ``single-element'' characters and 150$\times$2 compound characters covering all 150 radicals, to create a small dataset $S_D$ totally including 750 training samples.

\begin{figure*}[htb]
\centering
\setlength{\abovecaptionskip}{-3pt}
\hspace{-80pt}
\subfigure{
\begin{minipage}{0.8\textwidth}{
\begin{minipage}{0.10\textwidth}\textit{Source}\end{minipage}
\hspace{10pt}
\begin{minipage}{0.50\textwidth}
\includegraphics[width=1.8\textwidth]{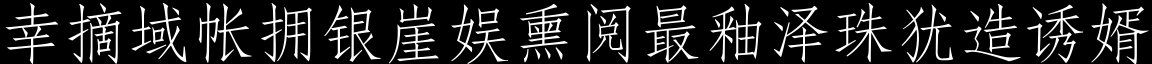}
\end{minipage}
\hspace{180pt}
\begin{minipage}{0.008\textwidth}\textit{RMSE}
\end{minipage}\\
}
\end{minipage}
}

\vspace{-5pt}
\ruleh
\hspace{-80pt}
\subfigure{
\begin{minipage}{0.8\textwidth}{
\begin{minipage}{0.10\textwidth}\textit{EMD few-shot}\end{minipage}
\hspace{10pt}
\begin{minipage}{0.50\textwidth}
\includegraphics[width=1.8\textwidth]{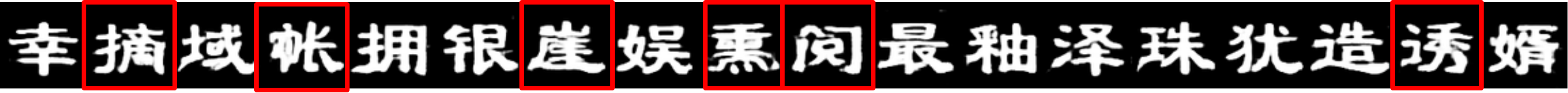}
\end{minipage}
\hspace{180pt}
\begin{minipage}{0.008\textwidth}\textit{4.888}
\end{minipage}\\
\begin{minipage}{0.10\textwidth}\textit{EMD finetune}\end{minipage}
\hspace{10pt}
\begin{minipage}{0.50\textwidth}
\includegraphics[width=1.8\textwidth]{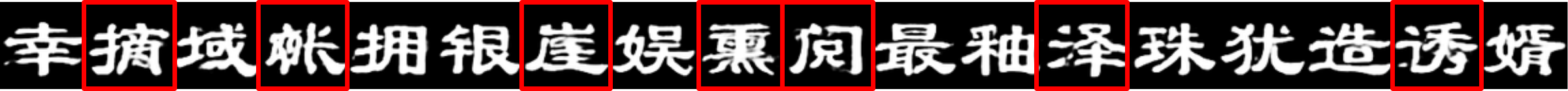}
\end{minipage}
\hspace{180pt}
\begin{minipage}{0.008\textwidth}\textit{4.964}
\end{minipage}\\
\begin{minipage}{0.10\textwidth}\textit{HAN}\end{minipage}
\hspace{10pt}
\begin{minipage}{0.50\textwidth}
\includegraphics[width=1.8\textwidth]{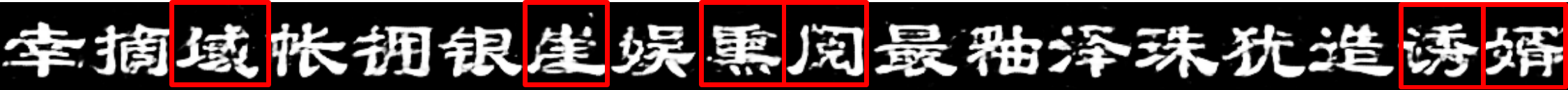}
\end{minipage}
\hspace{180pt}
\begin{minipage}{0.008\textwidth}\textit{4.844}
\end{minipage}\\
\begin{minipage}{0.10\textwidth}\textbf{\textit{Ours}}\end{minipage}
\hspace{10pt}
\begin{minipage}{0.50\textwidth}
\includegraphics[width=1.8\textwidth]{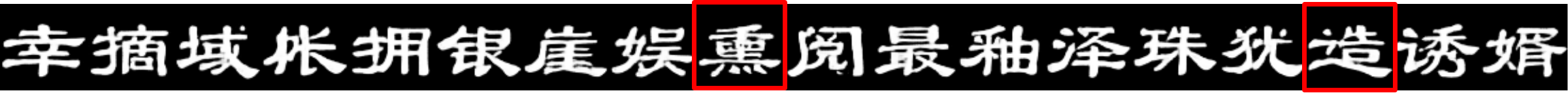}
\end{minipage}
\hspace{180pt}
\begin{minipage}{0.008\textwidth}\textit{\textbf{4.767}}
\end{minipage}\\
\begin{minipage}{0.10\textwidth}\textit{Target}\end{minipage}
\hspace{10pt}
\begin{minipage}{0.50\textwidth}
\includegraphics[width=1.8\textwidth]{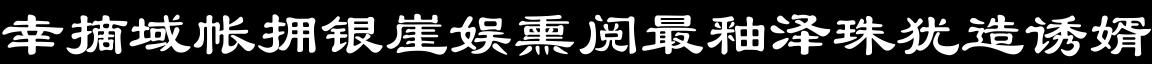}
\end{minipage}
\hspace{180pt}
\begin{minipage}{0.008\textwidth}\textit{}
\end{minipage}\\
}
\end{minipage}
}

\vspace{-5pt}
\ruleh
\hspace{-80pt}
\subfigure{
\begin{minipage}{0.8\textwidth}{
\begin{minipage}{0.10\textwidth}\textit{EMD few-shot}\end{minipage}
\hspace{10pt}
\begin{minipage}{0.50\textwidth}
\includegraphics[width=1.8\textwidth]{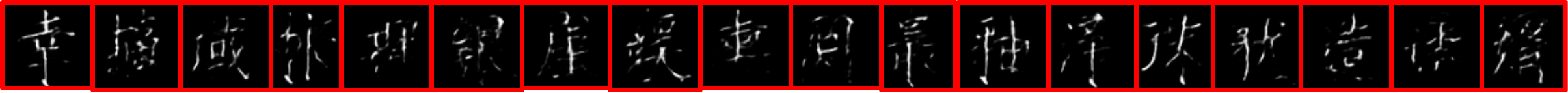}
\end{minipage}
\hspace{180pt}
\begin{minipage}{0.008\textwidth}\textit{4.965}
\end{minipage}\\
\begin{minipage}{0.10\textwidth}\textit{EMD finetune}\end{minipage}
\hspace{10pt}
\begin{minipage}{0.50\textwidth}
\includegraphics[width=1.8\textwidth]{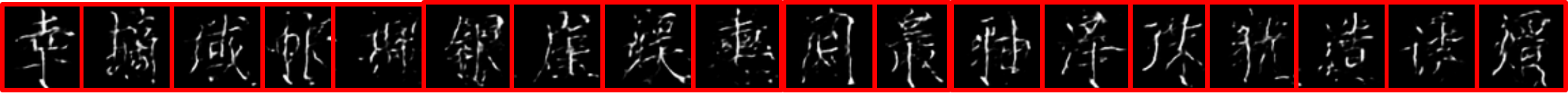}
\end{minipage}
\hspace{180pt}
\begin{minipage}{0.008\textwidth}\textit{4.999}
\end{minipage}\\
\begin{minipage}{0.10\textwidth}\textit{HAN}\end{minipage}
\hspace{10pt}
\begin{minipage}{0.50\textwidth}
\includegraphics[width=1.8\textwidth]{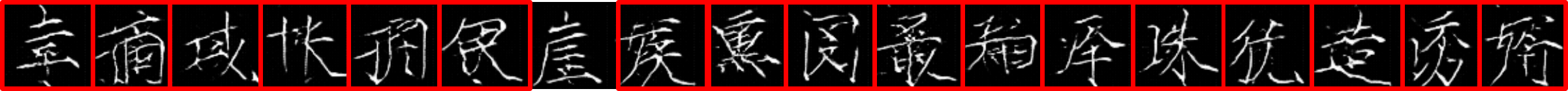}
\end{minipage}
\hspace{180pt}
\begin{minipage}{0.008\textwidth}\textit{4.886}
\end{minipage}\\
\begin{minipage}{0.10\textwidth}\textbf{\textit{Ours}}\end{minipage}
\hspace{10pt}
\begin{minipage}{0.50\textwidth}
\includegraphics[width=1.8\textwidth]{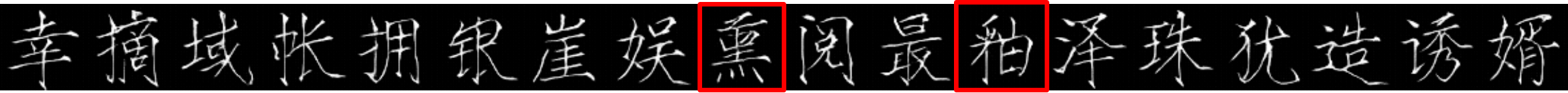}
\end{minipage}
\hspace{180pt}
\begin{minipage}{0.008\textwidth}\textit{\textbf{4.643}}
\end{minipage}\\
\begin{minipage}{0.10\textwidth}\textit{Target}\end{minipage}
\hspace{10pt}
\begin{minipage}{0.50\textwidth}
\includegraphics[width=1.8\textwidth]{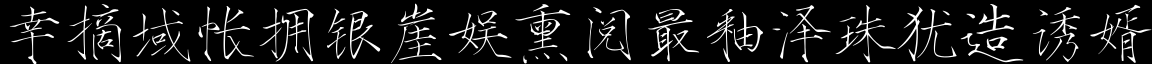}
\end{minipage}
\hspace{180pt}
\begin{minipage}{0.008\textwidth}\textit{}
\end{minipage}\\
}
\end{minipage}
}
 
\vspace{-5pt}
\ruleh
\hspace{-80pt}
\subfigure{
\begin{minipage}{0.8\textwidth}{
\begin{minipage}{0.10\textwidth}\textit{EMD few-shot}\end{minipage}
\hspace{10pt}
\begin{minipage}{0.50\textwidth}
\includegraphics[width=1.8\textwidth]{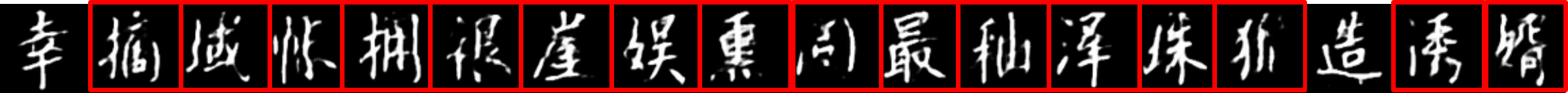}
\end{minipage}
\hspace{180pt}
\begin{minipage}{0.008\textwidth}\textit{5.095}
\end{minipage}\\
\begin{minipage}{0.10\textwidth}\textit{EMD finetune}\end{minipage}
\hspace{10pt}
\begin{minipage}{0.50\textwidth}
\includegraphics[width=1.8\textwidth]{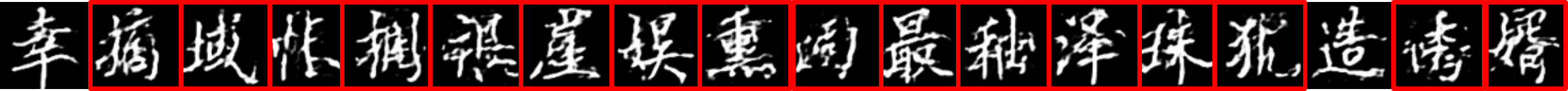}
\end{minipage}
\hspace{180pt}
\begin{minipage}{0.008\textwidth}\textit{5.525}
\end{minipage}\\
\begin{minipage}{0.10\textwidth}\textit{HAN}\end{minipage}
\hspace{10pt}
\begin{minipage}{0.50\textwidth}
\includegraphics[width=1.8\textwidth]{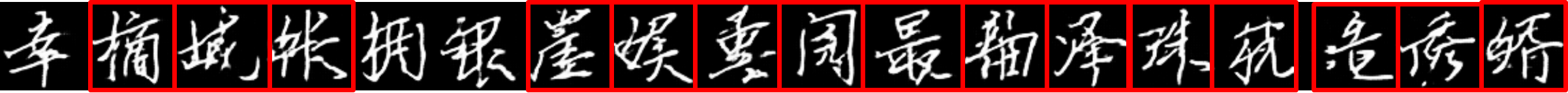}
\end{minipage}
\hspace{180pt}
\begin{minipage}{0.008\textwidth}\textit{5.113}
\end{minipage}\\
\begin{minipage}{0.10\textwidth}\textbf{\textit{Ours}}\end{minipage}
\hspace{10pt}
\begin{minipage}{0.50\textwidth}
\includegraphics[width=1.8\textwidth]{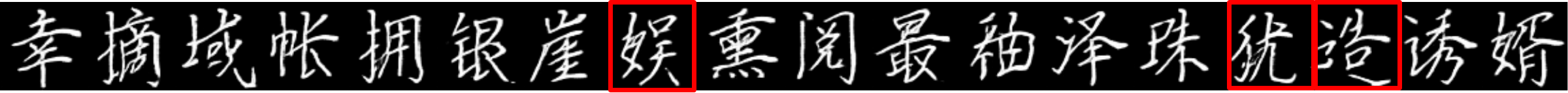}
\end{minipage}
\hspace{180pt}
\begin{minipage}{0.008\textwidth}\textit{\textbf{5.023}}
\end{minipage}\\
\begin{minipage}{0.10\textwidth}\textit{Target}\end{minipage}
\hspace{10pt}
\begin{minipage}{0.50\textwidth}
\includegraphics[width=1.8\textwidth]{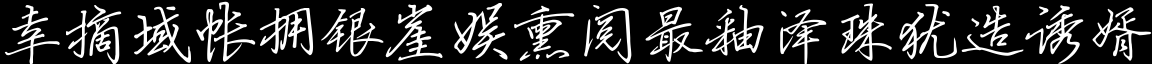}
\end{minipage}
\hspace{180pt}
\begin{minipage}{0.008\textwidth}\textit{}
\end{minipage}\\
}
\end{minipage}
}
\caption{Performance of transferring \textit{Fang Song} font (source) to other fonts (target) including \textbf{printed} (1st row) and \textbf{handwritten} fonts (2-3 rows). Characters in red boxes are some failure samples. Both qualitative results and quantitative evaluation (RMSE) demonstrate our model performs better than baselines.}
\label{fig:all_perfomance}
\end{figure*}

\begin{figure*}[htb]
\centering
\setlength{\abovecaptionskip}{-3pt}
\ruleh
\hspace{70pt}
\subfigure{
\begin{minipage}{0.2\textwidth}\textit{}\centering\end{minipage}
\hspace{-5pt}
\begin{minipage}{0.26\textwidth}\textit{Font1}\centering\end{minipage}
\begin{minipage}{0.26\textwidth}\textit{Font2}\centering\end{minipage}
\begin{minipage}{0.26\textwidth}\textit{Font3}\centering\end{minipage}
}

\vspace{-8pt}
\ruleh
\hspace{70pt}
\subfigure{
\begin{minipage}{0.2\textwidth}\textit{source}\centering\end{minipage}
\hspace{-5pt}
\begin{minipage}{0.26\textwidth}
\includegraphics[width=\textwidth]{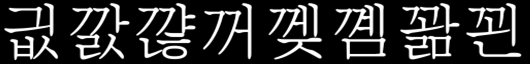}
\end{minipage}
\begin{minipage}{0.26\textwidth}
\includegraphics[width=\textwidth]{korean_src.png}
\end{minipage}
\begin{minipage}{0.26\textwidth}
\includegraphics[width=\textwidth]{korean_src.png}
\end{minipage}
}

\vspace{-8pt}
\hspace{70pt}
\subfigure{
\begin{minipage}{0.2\textwidth}\textit{result}\centering\end{minipage}
\hspace{-5pt}
\begin{minipage}{0.26\textwidth}
\includegraphics[width=\textwidth]{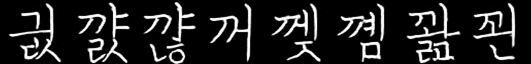}
\end{minipage}
\begin{minipage}{0.26\textwidth}
\includegraphics[width=\textwidth]{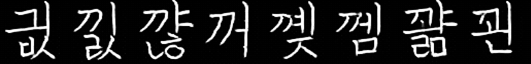}
\end{minipage}
\begin{minipage}{0.26\textwidth}
\includegraphics[width=\textwidth]{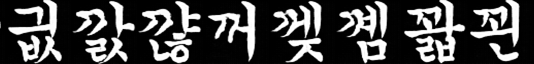}
\end{minipage}
}

\vspace{-8pt}
\hspace{70pt}
\subfigure{
\begin{minipage}{0.2\textwidth}\textit{target}\centering\end{minipage}
\hspace{-5pt}
\begin{minipage}{0.26\textwidth}
\includegraphics[width=\textwidth]{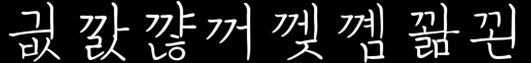}
\end{minipage}
\begin{minipage}{0.26\textwidth}
\includegraphics[width=\textwidth]{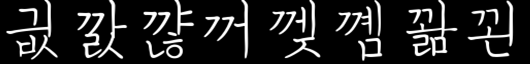}
\end{minipage}
\begin{minipage}{0.26\textwidth}
\includegraphics[width=\textwidth]{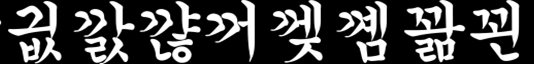}
\end{minipage}
}

\ruleh
\caption{The generated results of Korean fonts prove that our method can be transplanted to any languages with ``content reuse" phenomenon.}
\label{fig:korean}
\end{figure*}
\vspace{-2mm}
\subsection{Comparison with Baseline Methods}
\vspace{-1mm}
\label{sec: Comparison-with-Baseline-Methods}
Two state-of-the-art Chinese font synthesis methods we compared with are HAN and EMD. HAN\cite{changchinese} is especially proposed for handwritten fonts synthesis. It proposes hierarchical adversarial discriminator to improve the generated performance. Experiment shows it relies on about 2500 paired training samples to achieve good performance; EMD\cite{zhang2018separating} can achieve style transfer from a few samples by disentangling the style/content representation of fonts, while it relies on a large font library for training and performs not well on handwritten fonts.
\vspace{-3mm}
\subsubsection{Performance on Small Dataset $S_D$}
We first experimentally compare our method with HAN\cite{changchinese} and EMD\cite{zhang2018separating} under the selected small dataset $S_D$(see Fig. \ref{fig:all_perfomance}). Specially, EMD (few-shot) denotes we generate characters using a few reference inputs, just like what the paper has done. However, for fair comparison, EMD (fine-tune) denotes we further fine tune EMD with $S_D$. For HAN, we directly train it on $S_D$.
We specially choose both \emph{\bf printed-style} and \emph{\bf handwritten-style} fonts to fairly demonstrate our model generally outperforms baselines. According to Fig. \ref{fig:all_perfomance}, our model slightly outperforms baseline methods on printed-style fonts (1st row) since printed-style fonts are always featured with regular structure and wide strokes, which makes the model take no advantages of proposed \textit{collaborative stroke refinement}.
However, our method achieves impressive results on handwritten fonts featured with thin strokes(2nd row) or even irregular structure (3rd rows). Compared with baselines, our model synthesizes more details without missing or overlapped strokes. While these defections happen in baseline methods, we can barely recognize some synthesized characters of them. More experimental results are displayed in appendix.

\begin{figure*}[htb]
\centering
\setlength{\abovecaptionskip}{-3pt}
\hspace{-10pt}
\subfigure{
\begin{minipage}{0.8\textwidth}{
\begin{minipage}{0.14\textwidth}\textit{Source}\end{minipage}
\begin{minipage}{0.50\textwidth}
\includegraphics[width=1.8\textwidth]{source_18.png}
\end{minipage}\\
}
\end{minipage}
}
\vspace{-7pt}
\ruleh
\hspace{-15pt}
\subfigure{
\begin{minipage}{0.8\textwidth}{
\begin{minipage}{0.14\textwidth}\textit{EMD finetune 750}\end{minipage}
\begin{minipage}{0.50\textwidth}
\includegraphics[width=1.8\textwidth]{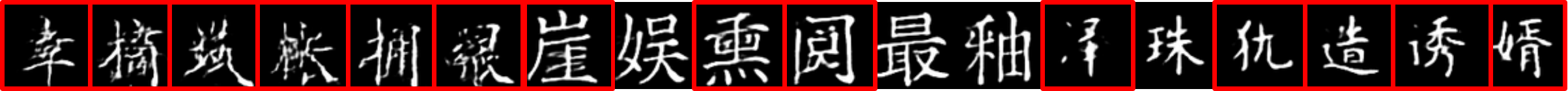}
\end{minipage}\\
\begin{minipage}{0.14\textwidth}\textit{HAN 750}\end{minipage}
\begin{minipage}{0.50\textwidth}
\includegraphics[width=1.8\textwidth]{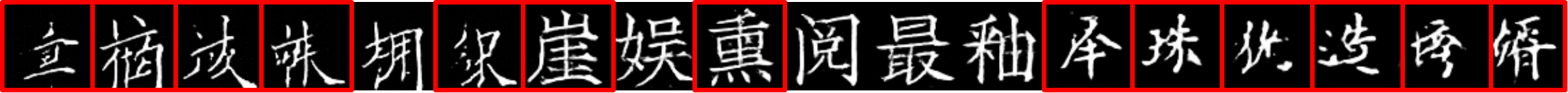}
\end{minipage}\\
\begin{minipage}{0.14\textwidth}{\color{blue}\textbf{\textit{Ours 750}}}\end{minipage}
\begin{minipage}{0.50\textwidth}
\includegraphics[width=1.8\textwidth]{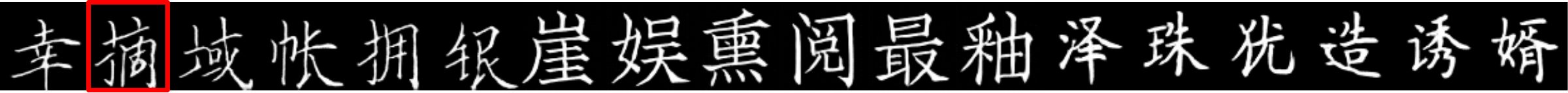}
\end{minipage}\\
}
\end{minipage}
}

\vspace{-7pt}
\ruleh
\hspace{-15pt}
\subfigure{
\begin{minipage}{0.8\textwidth}{
\begin{minipage}{0.14\textwidth}\textit{EMD finetune 1550}\end{minipage}
\begin{minipage}{0.50\textwidth}
\includegraphics[width=1.8\textwidth]{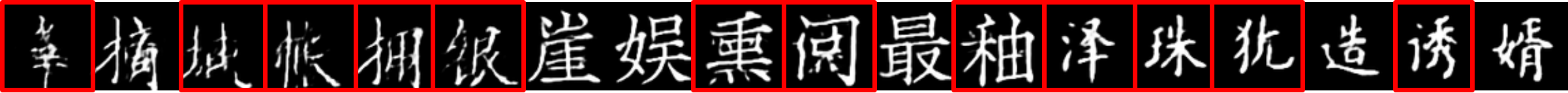}
\end{minipage}\\
\begin{minipage}{0.14\textwidth}\textit{HAN 1550}\end{minipage}
\begin{minipage}{0.50\textwidth}
\includegraphics[width=1.8\textwidth]{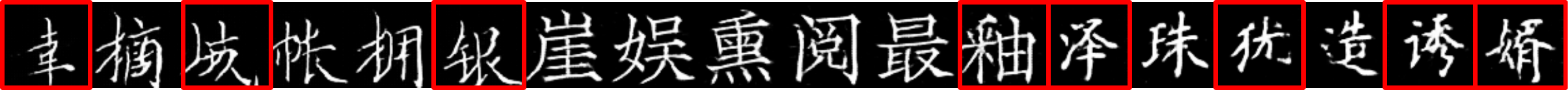}
\end{minipage}\\
\begin{minipage}{0.14\textwidth}\textbf{\textit{Ours 1550}}\end{minipage}
\begin{minipage}{0.50\textwidth}
\includegraphics[width=1.8\textwidth]{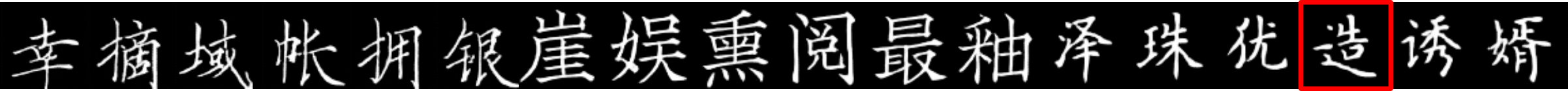}
\end{minipage}\\
}
\end{minipage}
}

\vspace{-7pt}
\ruleh
\hspace{-15pt}
\subfigure{
\begin{minipage}{0.8\textwidth}{
\begin{minipage}{0.14\textwidth}\textit{EMD finetune 2550}\end{minipage}
\begin{minipage}{0.50\textwidth}
\includegraphics[width=1.8\textwidth]{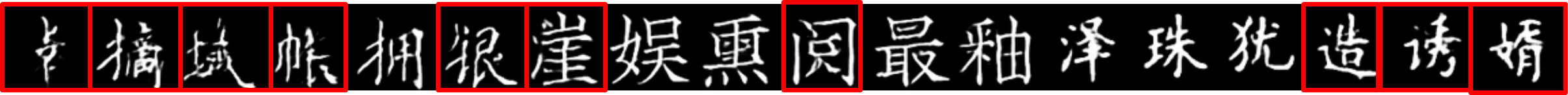}
\end{minipage}\\
\begin{minipage}{0.14\textwidth}\color{blue}\textit{HAN 2550}\end{minipage}
\begin{minipage}{0.50\textwidth}
\includegraphics[width=1.8\textwidth]{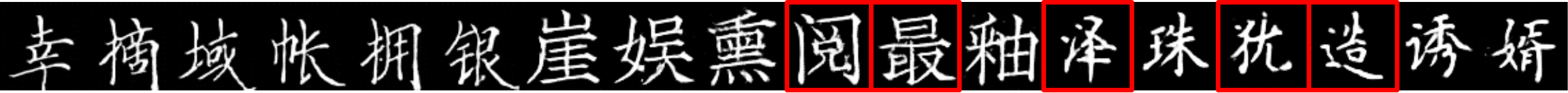}
\end{minipage}\\
\begin{minipage}{0.14\textwidth}\textbf{\textit{Ours 2550}}\end{minipage}
\begin{minipage}{0.50\textwidth}
\includegraphics[width=1.8\textwidth]{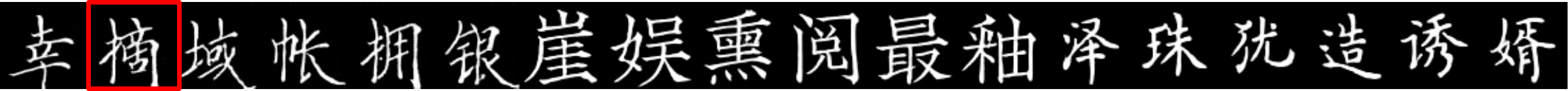}
\end{minipage}\\
}
\end{minipage}
}

\vspace{-7pt}
\ruleh
\hspace{-15pt}
\subfigure{
\begin{minipage}{0.8\textwidth}{
\begin{minipage}{0.14\textwidth}\textbf{\textit{Target}}\end{minipage}
\begin{minipage}{0.50\textwidth}
\includegraphics[width=1.8\textwidth]{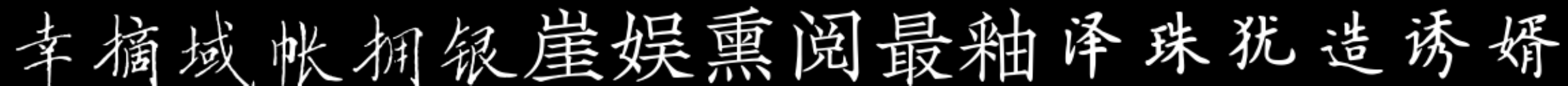}
\end{minipage}\\
}
\end{minipage}
}

\caption{Performance comparison on three handwritten font synthesis tasks  with increasing size of training set. The comparison between {\color{blue}\textit{Ours 750}} and {\color{blue}\textit{HAN 2550}} demonstrate that our method achieve the equal even better performance with much less training set.}
\label{fig: training size}
\end{figure*}
\begin{figure*}[htb]
\centering
\setlength{\abovecaptionskip}{-3pt}

\ruleh
\hspace{70pt}
\subfigure{
\begin{minipage}{0.2\textwidth}\textit{}\centering\end{minipage}
\hspace{-5pt}
\begin{minipage}{0.26\textwidth}\textit{Font1}\centering\end{minipage}
\begin{minipage}{0.26\textwidth}\textit{Font2}\centering\end{minipage}
\begin{minipage}{0.26\textwidth}\textit{Font3}\centering\end{minipage}
}

\vspace{-6pt}
\ruleh
\hspace{70pt}
\subfigure{
\begin{minipage}{0.2\textwidth}\textit{source}\centering\end{minipage}
\hspace{-5pt}
\begin{minipage}{0.26\textwidth}
\includegraphics[width=\textwidth]{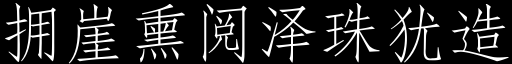}
\end{minipage}
\begin{minipage}{0.26\textwidth}
\includegraphics[width=\textwidth]{source_8.png}
\end{minipage}
\begin{minipage}{0.26\textwidth}
\includegraphics[width=\textwidth]{source_8.png}
\end{minipage}
}

\vspace{-10pt}
\hspace{70pt}
\subfigure{
\begin{minipage}{0.2\textwidth}\textit{w/o pre\_deformation}\centering\end{minipage}
\hspace{-5pt}
\begin{minipage}{0.26\textwidth}
\includegraphics[width=\textwidth]{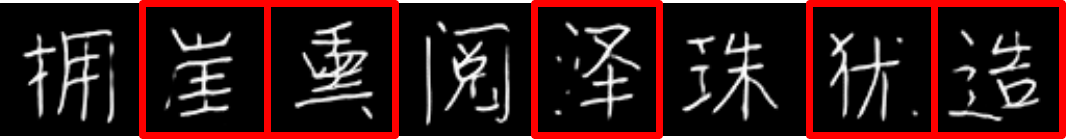}
\end{minipage}
\begin{minipage}{0.26\textwidth}
\includegraphics[width=\textwidth]{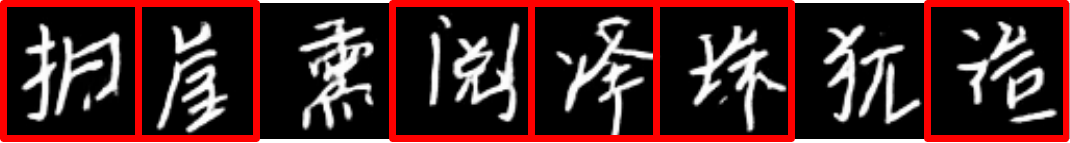}
\end{minipage}
\begin{minipage}{0.26\textwidth}
\includegraphics[width=\textwidth]{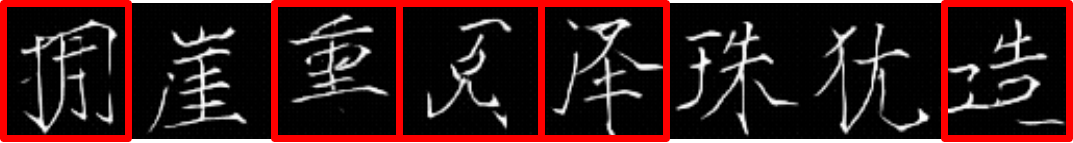}
\end{minipage}
}

\vspace{-10pt}
\hspace{70pt}
\subfigure{
\begin{minipage}{0.2\textwidth}\textit{w/o augmentation}\centering\end{minipage}
\hspace{-5pt}
\begin{minipage}{0.26\textwidth}
\includegraphics[width=\textwidth]{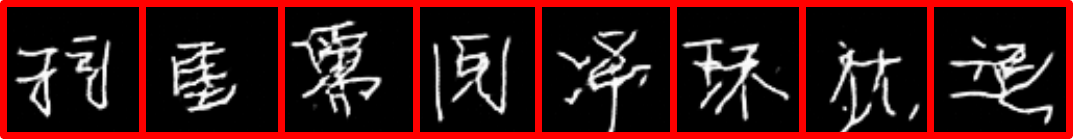}
\end{minipage}
\begin{minipage}{0.26\textwidth}
\includegraphics[width=\textwidth]{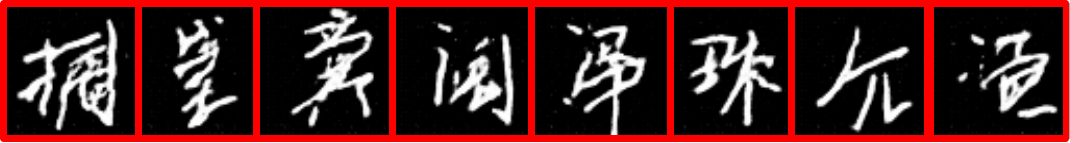}
\end{minipage}
\begin{minipage}{0.26\textwidth}
\includegraphics[width=\textwidth]{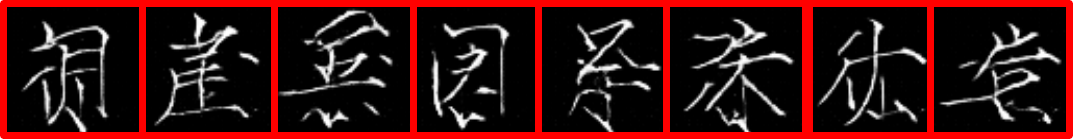}
\end{minipage}
}

\vspace{-10pt}
\hspace{70pt}
\subfigure{
\begin{minipage}{0.2\textwidth}\textit{w/o refinement}\centering\end{minipage}
\hspace{-5pt}
\begin{minipage}{0.26\textwidth}
\includegraphics[width=\textwidth]{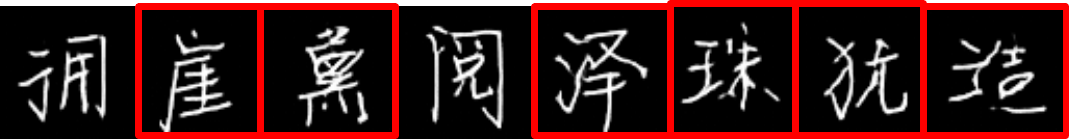}
\end{minipage}
\begin{minipage}{0.26\textwidth}
\includegraphics[width=\textwidth]{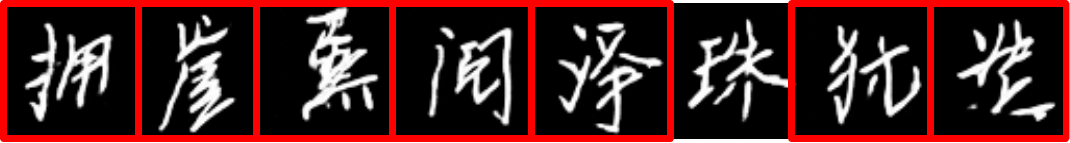}
\end{minipage}
\begin{minipage}{0.26\textwidth}
\includegraphics[width=\textwidth]{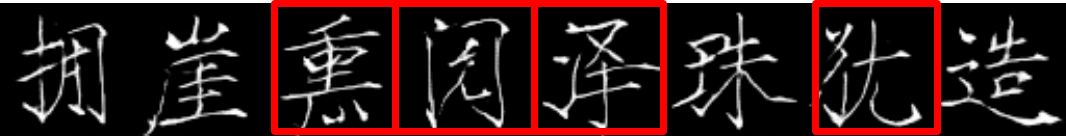}
\end{minipage}
}

\vspace{-10pt}
\hspace{70pt}
\subfigure{
\begin{minipage}{0.2\textwidth}\textit{w/o GAN}\centering\end{minipage}
\hspace{-5pt}
\begin{minipage}{0.26\textwidth}
\includegraphics[width=\textwidth]{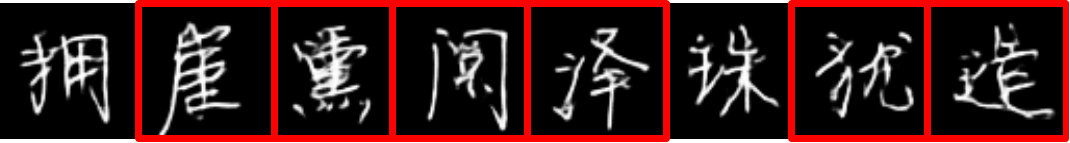}
\end{minipage}
\begin{minipage}{0.26\textwidth}
\includegraphics[width=\textwidth]{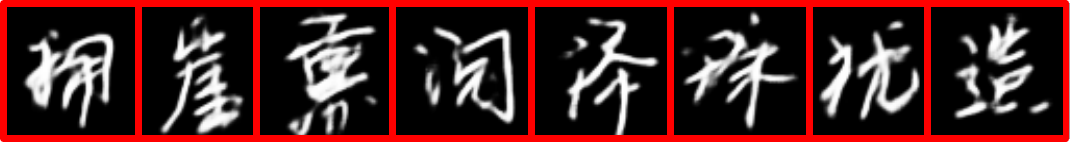}
\end{minipage}
\begin{minipage}{0.26\textwidth}
\includegraphics[width=\textwidth]{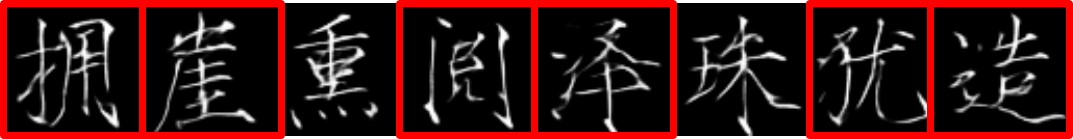}
\end{minipage}
}

\vspace{-10pt}
\hspace{70pt}
\subfigure{
\begin{minipage}{0.2\textwidth}\textit{\textbf{all}}\centering\end{minipage}
\hspace{-5pt}
\begin{minipage}{0.26\textwidth}
\includegraphics[width=\textwidth]{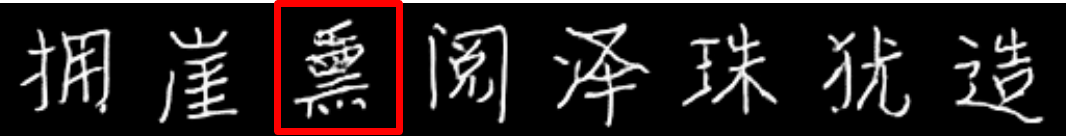}
\end{minipage}
\begin{minipage}{0.26\textwidth}
\includegraphics[width=\textwidth]{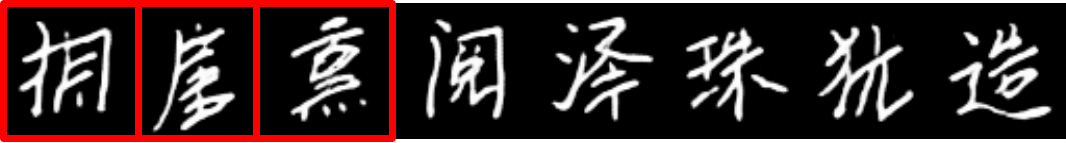}
\end{minipage}
\begin{minipage}{0.26\textwidth}
\includegraphics[width=\textwidth]{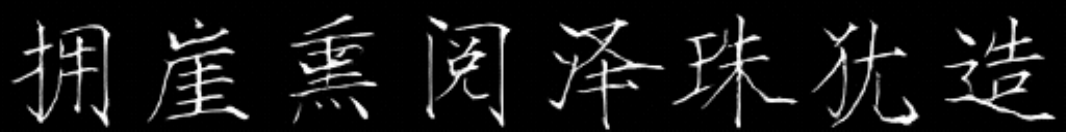}
\end{minipage}
}

\vspace{-10pt}
\hspace{70pt}
\subfigure{
\begin{minipage}{0.2\textwidth}\textit{target}\centering\end{minipage}
\hspace{-5pt}
\begin{minipage}{0.26\textwidth}
\includegraphics[width=\textwidth]{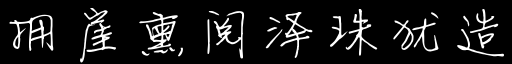}
\end{minipage}
\begin{minipage}{0.26\textwidth}
\includegraphics[width=\textwidth]{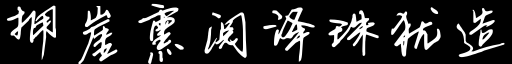}
\end{minipage}
\begin{minipage}{0.26\textwidth}
\includegraphics[width=\textwidth]{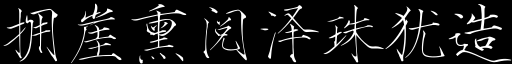}
\end{minipage}
}

\ruleh
\caption{Ablation comparison: for each example font, we show predictions without adaptive pre-deformation, online zoom-augmentation, stroke refinement, influence of GAN-related losses predictions and last with all the methods. The results are most similar to the target while the performance under other ablation setting shows various defections.}
\label{fig:ablation}
\end{figure*}



\vspace{-3mm}
\subsubsection{Performance on Different Training-set Size}
We further change the size of training set to 1550 and 2550 by randomly adding additional samples to $S_D$. we compare our model with baselines under various sizes of training set on handwritten font synthesis tasks (see Fig. \ref{fig: training size}). 

Besides RMSE, to rationally measure the fidelity of synthesized characters, we conduct a user Study (see Fig. \ref{fig: table}). $100$ volunteers are invited to subjectively rate the synthesized characters from score 1 to 5, where 1 is the worst and 5 is the best. The user study results show that the performance of all methods is improved with the increase of the training set. However, when the size is larger than 1550, the increasing trend of our method stops and the score begins to float up and down. Thus we conclude that 1550 training samples have completely covered radicals/single-element characters with different shapes so that more training samples will not bring more improvement. Additionally, User Study result demonstrates that when the size of training set is 750, our method achieves equal even higher subjective score compared with HAN trained by 2550 characters.

\vspace{-3mm}
\subsection{Ablation Study}
\textbf{Effect of Adaptive Pre-deformation} 
We conduct experiments by removing adaptive pre-deformation or not.
As shown in the second row of Fig. \ref{fig:ablation}, some strokes are missing compared with the ground truth, while the results of ours are significantly better, which means pre-deformation guarantees that the generated strokes are complete. 
When absolute locations of a certain stroke between the source character and the target character are seriously discrepant, the transfer network may be confused about whether this stroke should be abandoned or be mapped to another position. Our adaptive pre-deformation roughly align the strokes essentially relieves the transfer network of learning the mapping of stroke location.

\textbf{Effect of online zoom-augmentation}
The results after removing online augmentation module are shown in the third row of Fig. \ref{fig:ablation} from which we can see the generated strokes are so disordered that we even cannot recognize the characters. 
Without zoom-augmentation, the model is actually trained with only 750 paired samples, which may lead to serious over-fitting. Zoom-augmentation boosts the model implicitly learn the shape diversity and location diversity of one character. Besides, it also models common structure information by these vertically or horizontally translated characters. So our method can reconstruct correct topological structure of characters.  

\textbf{Effect of Stroke Refinement}
We disconnect the data flow from \textit{refine branch} to \textit{dominating branch} for analyzing the effect of stroke refinement. Comparison results illustrated in Fig. \ref{fig:ablation} demonstrate that characters generated with stroke refinement strategy have more continuous structures without missing or broken strokes, while characters generated without the participation of \textit{stroke refinement} present seriously deviant and even broken strokes, especially on cursive handwritten fonts (\textit{Font} 1 and \textit{Font} 2). These phenomena proves the effectiveness of stroke refinement.

\textbf{Effect of GAN}
We last remove the influence of two discriminators in the model by setting GAN-related loss terms, Eq. \ref{eq: eq7} and Eq. \ref{eq: eq9}, to zero. The results in Fig. \ref{fig:ablation} generally shows a little blurry.


\subsection{Extension Study}
To show the portability of our approach, we experiment Korean fonts. Fig. \ref{fig:korean} shows that the syntheses are impressively similar to the targets, validating our model can be applied to other font generation tasks, once they have complicated handwriting style and content reuse phenomenon.


\section{Conclusions}
We propose an end-to-end model to synthesize handwritten Chinese font with only 750 training samples. The whole model includes three main novelties: collaborative stroke refinement, handling the thin issue, 
online zoom-augmentation, exloiting the content-reuse phenomenon, and  adaptive pre-deformation, augmenting training data. We perceptually and quantitatively evaluate our model and the experimental results validate the effectiveness of our model. 


{\small
\bibliographystyle{ieee}
\bibliography{main}
}

\newpage

\section{Appendix}
\subsection{Appendix \uppercase\expandafter{\romannumeral1}}

\begin{figure*}[ht]
\subsection{Appendix \uppercase\expandafter{\romannumeral1}}
\centering
\setlength{\abovecaptionskip}{-3pt}
\hspace{-26pt}
\subfigure{
\begin{minipage}{0.8\textwidth}{
\begin{minipage}{0.18\textwidth}\textit{Source}\centering\end{minipage}
\begin{minipage}{0.50\textwidth}
\includegraphics[width=1.8\textwidth]{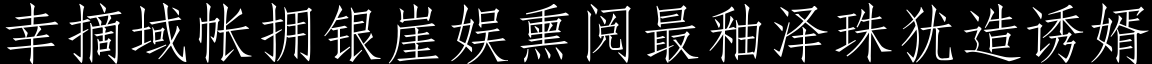}
\end{minipage}\\
}
\end{minipage}
}
\vspace{-7pt}
\ruleh
\hspace{-30pt}
\subfigure{
\begin{minipage}{0.8\textwidth}{
\begin{minipage}{0.18\textwidth}\textit{$\widehat{y}$ w/o refinement}\centering\end{minipage}
\begin{minipage}{0.50\textwidth}
\includegraphics[width=1.8\textwidth]{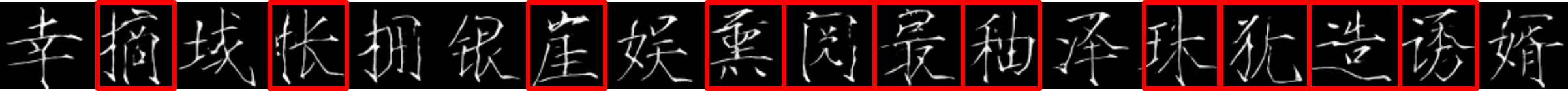}
\end{minipage}\\
\begin{minipage}{0.18\textwidth}\textit{$\widehat{b(y)}$}\centering\end{minipage}
\begin{minipage}{0.50\textwidth}
\includegraphics[width=1.8\textwidth]{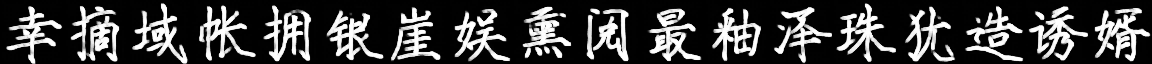}
\end{minipage}\\
\begin{minipage}{0.18\textwidth}\textit{$g(\widehat{b(y)})$}\centering\end{minipage}
\begin{minipage}{0.50\textwidth}
\includegraphics[width=1.8\textwidth]{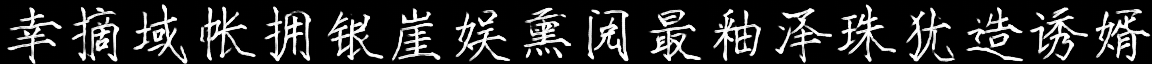}
\end{minipage}\\
\begin{minipage}{0.18\textwidth}\textbf{\textit{$\widehat{y}$ our}}\centering\end{minipage}
\begin{minipage}{0.50\textwidth}
\includegraphics[width=1.8\textwidth]{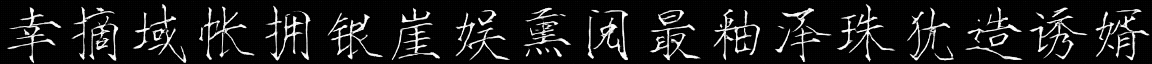}
\end{minipage}\\
\begin{minipage}{0.18\textwidth}\textit{target}\centering\end{minipage}
\begin{minipage}{0.50\textwidth}
\includegraphics[width=1.8\textwidth]{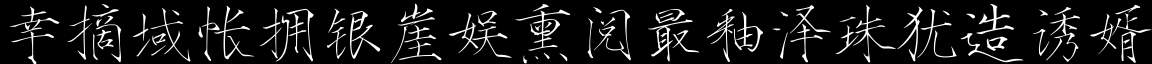}
\end{minipage}\\
}
\end{minipage}
}

\vspace{-7pt}
\ruleh
\hspace{-30pt}
\subfigure{
\begin{minipage}{0.8\textwidth}{
\begin{minipage}{0.18\textwidth}\textit{$\widehat{y}$ w/o refinement}\centering\end{minipage}
\begin{minipage}{0.50\textwidth}
\includegraphics[width=1.8\textwidth]{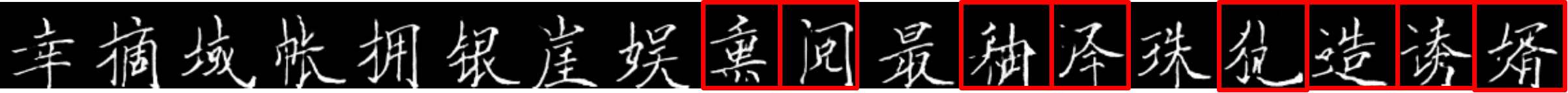}
\end{minipage}\\
\begin{minipage}{0.18\textwidth}\textit{$\widehat{b(y)}$}\centering\end{minipage}
\begin{minipage}{0.50\textwidth}
\includegraphics[width=1.8\textwidth]{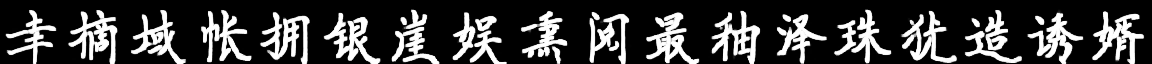}
\end{minipage}\\
\begin{minipage}{0.18\textwidth}\textit{$g(\widehat{b(y)})$}\centering\end{minipage}
\begin{minipage}{0.50\textwidth}
\includegraphics[width=1.8\textwidth]{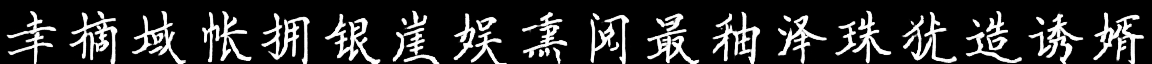}
\end{minipage}\\
\begin{minipage}{0.18\textwidth}\textbf{\textit{\textit{$\widehat{b(y)}$ our}}}\centering\end{minipage}
\begin{minipage}{0.50\textwidth}
\includegraphics[width=1.8\textwidth]{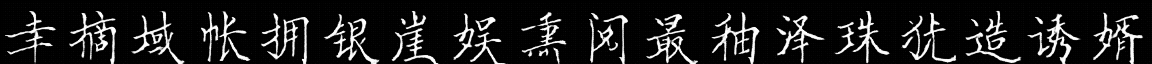}
\end{minipage}\\
\begin{minipage}{0.18\textwidth}\textit{target}\centering\end{minipage}
\begin{minipage}{0.50\textwidth}
\includegraphics[width=1.8\textwidth]{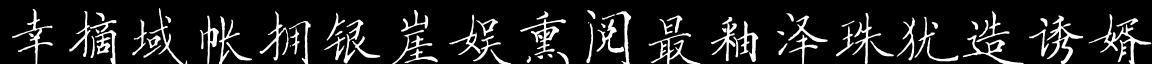}
\end{minipage}\\
}
\end{minipage}
}

\vspace{-7pt}
\ruleh
\hspace{-30pt}
\subfigure{
\begin{minipage}{0.8\textwidth}{
\begin{minipage}{0.18\textwidth}\textit{$\widehat{y}$ w/o refinement}\centering\end{minipage}
\begin{minipage}{0.50\textwidth}
\includegraphics[width=1.8\textwidth]{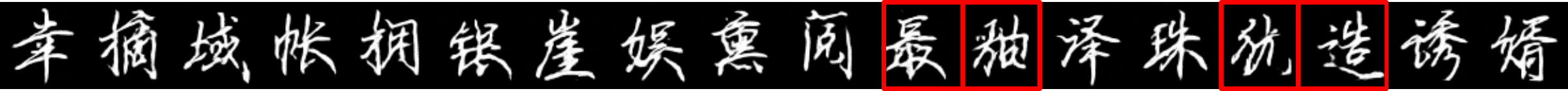}
\end{minipage}\\
\begin{minipage}{0.18\textwidth}\textit{$\widehat{b(y)}$}\centering\end{minipage}
\begin{minipage}{0.50\textwidth}
\includegraphics[width=1.8\textwidth]{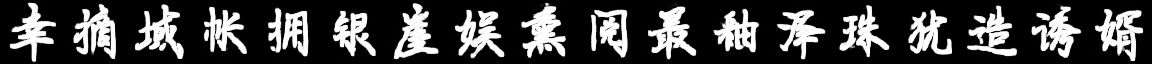}
\end{minipage}\\
\begin{minipage}{0.18\textwidth}\textit{$g(\widehat{b(y)})$}\centering\end{minipage}
\begin{minipage}{0.50\textwidth}
\includegraphics[width=1.8\textwidth]{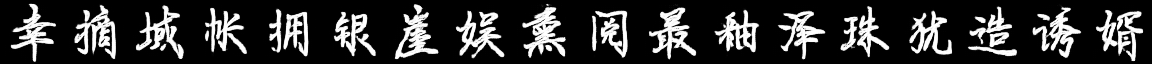}
\end{minipage}\\
\begin{minipage}{0.18\textwidth}\textbf{\textit{\textit{$\widehat{y}$ our}}}\centering\end{minipage}
\begin{minipage}{0.50\textwidth}
\includegraphics[width=1.8\textwidth]{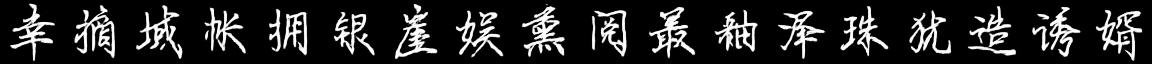}
\end{minipage}\\
\begin{minipage}{0.18\textwidth}\textit{target}\centering\end{minipage}
\begin{minipage}{0.50\textwidth}
\includegraphics[width=1.8\textwidth]{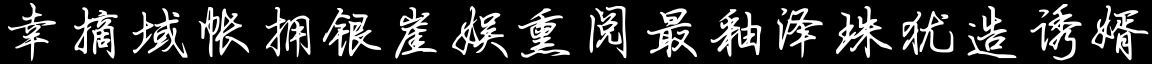}
\end{minipage}\\
}
\end{minipage}
}

\vspace{-7pt}
\ruleh
\hspace{-30pt}
\subfigure{
\begin{minipage}{0.8\textwidth}{
\begin{minipage}{0.18\textwidth}\textit{$\widehat{y}$ w/o refinement}\centering\end{minipage}
\begin{minipage}{0.50\textwidth}
\includegraphics[width=1.8\textwidth]{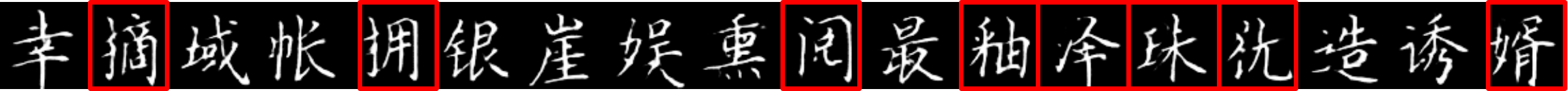}
\end{minipage}\\
\begin{minipage}{0.18\textwidth}\textit{$\widehat{b(y)}$}\centering\end{minipage}
\begin{minipage}{0.50\textwidth}
\includegraphics[width=1.8\textwidth]{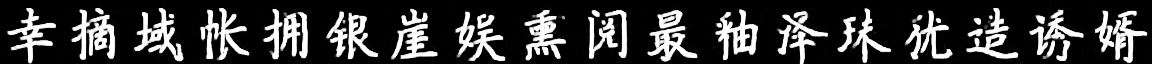}
\end{minipage}\\
\begin{minipage}{0.18\textwidth}\textit{$g(\widehat{b(y)})$}\centering\end{minipage}
\begin{minipage}{0.50\textwidth}
\includegraphics[width=1.8\textwidth]{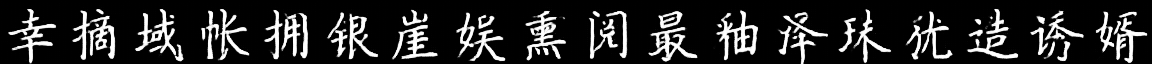}
\end{minipage}\\
\begin{minipage}{0.18\textwidth}\textbf{\textit{\textit{$\widehat{y}$ our}}}\centering\end{minipage}
\begin{minipage}{0.50\textwidth}
\includegraphics[width=1.8\textwidth]{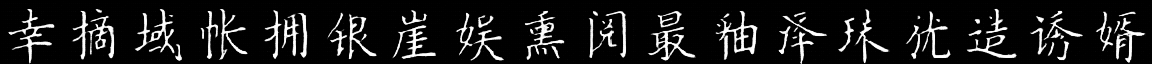}
\end{minipage}\\
\begin{minipage}{0.18\textwidth}\textit{target}\centering\end{minipage}
\begin{minipage}{0.50\textwidth}
\includegraphics[width=1.8\textwidth]{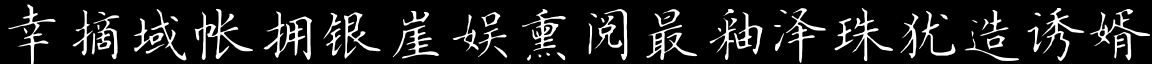}
\end{minipage}\\
}
\end{minipage}
}

\vspace{-7pt}
\ruleh
\hspace{-30pt}
\subfigure{
\begin{minipage}{0.8\textwidth}{
\begin{minipage}{0.18\textwidth}\textit{$\widehat{y}$ w/o refinement}\centering\end{minipage}
\begin{minipage}{0.50\textwidth}
\includegraphics[width=1.8\textwidth]{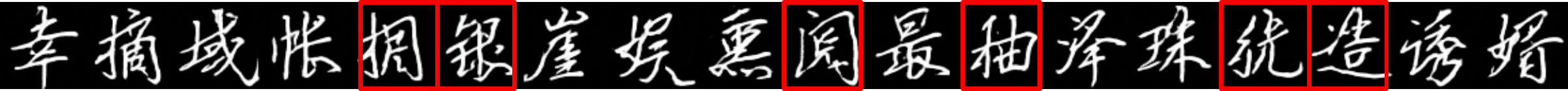}
\end{minipage}\\
\begin{minipage}{0.18\textwidth}\textit{$\widehat{b(y)}$}\centering\end{minipage}
\begin{minipage}{0.50\textwidth}
\includegraphics[width=1.8\textwidth]{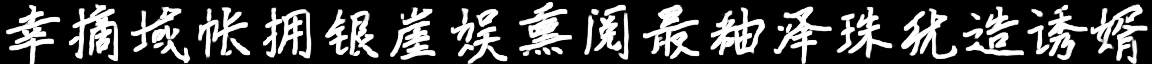}
\end{minipage}\\
\begin{minipage}{0.18\textwidth}\textit{$g(\widehat{b(y)})$}\centering\end{minipage}
\begin{minipage}{0.50\textwidth}
\includegraphics[width=1.8\textwidth]{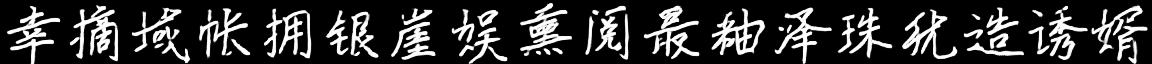}
\end{minipage}\\
\begin{minipage}{0.18\textwidth}\textbf{\textit{\textit{$\widehat{y}$} our}}\centering\end{minipage}
\begin{minipage}{0.50\textwidth}
\includegraphics[width=1.8\textwidth]{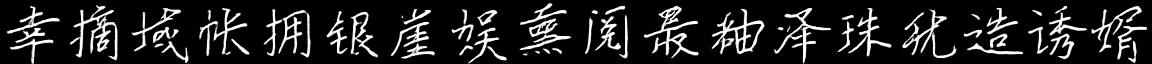}
\end{minipage}\\
\begin{minipage}{0.18\textwidth}\textit{target}\centering\end{minipage}
\begin{minipage}{0.50\textwidth}
\includegraphics[width=1.8\textwidth]{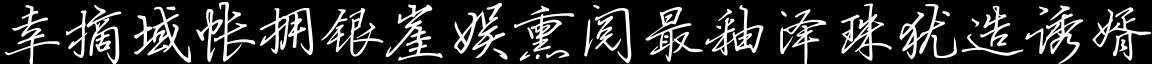}
\end{minipage}\\
}
\end{minipage}
}

\caption{Intermediate results of the model and the results of model without stroke refinement.  The characters in red boxes are seriously distorted because of removing refinement module. It proves that \textit{collaborative stroke refinement} strategy makes the strokes more continuous, smooth and sharp.}
\label{fig: 1}
\end{figure*}

As shown in Figure \ref{fig: 1}, we get the intermediate results $\widehat{b(y)}$ and $g(\widehat{b(y)})$ during the training process to analyze the pipeline of our model. Besides, we also conduct an experiment by removing the \textit{collaborative stroke refinement} to show the importance of this module. The results of our model are much more similar to targets than the model without stroke refinement. We mark the distorted characters with red bounding boxes. From the number of bounding boxes, we can learn that the first row performs worst because its strokes are thinnest. Removing the \textit{collaborative stroke refinement}, many strokes of generated characters are broken because of the low fault tolerance. But our model improve the fault tolerance by generating and refining $\widehat{b(y)}$. Figure \ref{fig: 1} illustrates the whole process of \textit{collaborative stroke refinement} and shows its importance.

\subsection{Appendix \uppercase\expandafter{\romannumeral2}}

\begin{figure*}
\subsection{Appendix \uppercase\expandafter{\romannumeral2}}
\centering
\setlength{\abovecaptionskip}{-3pt}

\vspace{-7pt}
\ruleh
\hspace{-90pt}
\subfigure{
\begin{minipage}{0.8\textwidth}{
\begin{minipage}{0.11\textwidth}\textit{Source1} \centering\end{minipage}
\begin{minipage}{0.50\textwidth}
\includegraphics[width=2.2\textwidth]{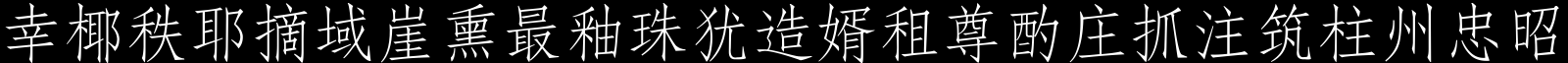}
\end{minipage}\\
\begin{minipage}{0.11\textwidth}\textit{font1} \centering\end{minipage}
\begin{minipage}{0.50\textwidth}
\includegraphics[width=2.2\textwidth]{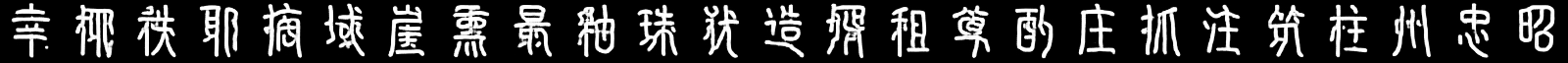}
\end{minipage}\\
\begin{minipage}{0.11\textwidth}\textit{target}\centering\end{minipage}
\begin{minipage}{0.50\textwidth}
\includegraphics[width=2.2\textwidth]{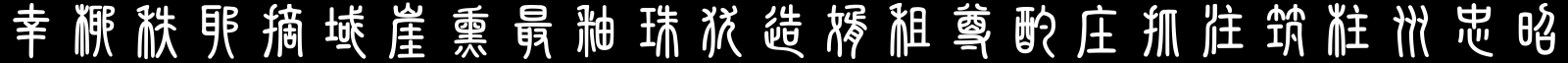}
\end{minipage}\\
}
\end{minipage}
}

\vspace{-7pt}
\ruleh
\hspace{-90pt}
\subfigure{
\begin{minipage}{0.8\textwidth}{
\begin{minipage}{0.11\textwidth}\textit{Source2} \centering\end{minipage}
\begin{minipage}{0.50\textwidth}
\includegraphics[width=2.2\textwidth]{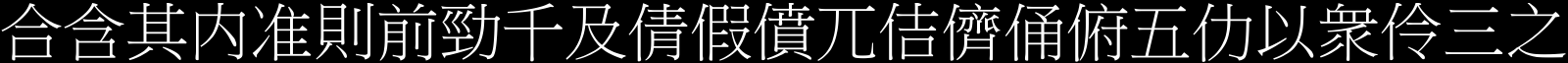}
\end{minipage}\\
\begin{minipage}{0.11\textwidth}\textit{font2}\centering\end{minipage}
\begin{minipage}{0.50\textwidth}
\includegraphics[width=2.2\textwidth]{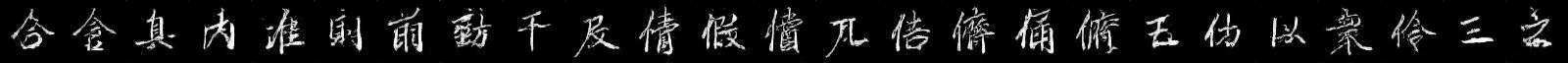}
\end{minipage}\\
\begin{minipage}{0.11\textwidth}\textit{target}\centering\end{minipage}
\begin{minipage}{0.50\textwidth}
\includegraphics[width=2.2\textwidth]{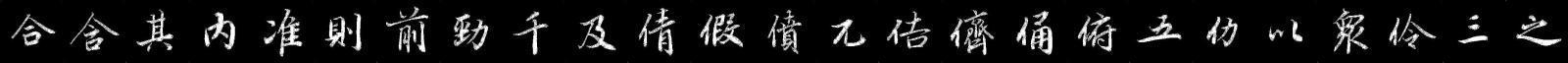}
\end{minipage}\\
}
\end{minipage}
}

\vspace{-7pt}
\ruleh

\caption{Results of ancient Chinese characters and calligraphy. Font 1 is a type of ancient Chinese characters named Xiao Zhuan, which has more distorted strokes than modern characters. Font 2 is a calligraphy duplicated from Rubbings written by a famous calligrapher in Tang Dynasty. The performance of our model on these two fonts are pretty good, which means our model can be applied in history and art research.}
\label{fig: 2}
\end{figure*}

As shown in Figure \ref{fig: 2}, we conduct extensive experiments on ancient Chinese characters and calligraphy. Font 1 is a type of ancient font named Xiao Zhuan and Font 2 is a calligraphy duplicated from rubbings written by a famous calligrapher in Tang Dynasty. The results reconstruct the overall glyph and structure of targets. The results can be applied to history and art research area.

\begin{figure*}
\centering
\setlength{\abovecaptionskip}{-3pt}
\hspace{-86pt}
\subfigure{
\begin{minipage}{0.8\textwidth}{
\begin{minipage}{0.18\textwidth}\textit{Source}\centering\end{minipage}
\begin{minipage}{0.50\textwidth}
\includegraphics[width=1.9\textwidth]{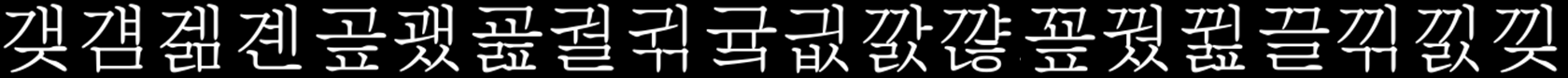}
\end{minipage}\\
}
\end{minipage}
}
\vspace{-7pt}
\ruleh
\hspace{-90pt}
\subfigure{
\begin{minipage}{0.8\textwidth}{
\begin{minipage}{0.18\textwidth}\textit{font1} \centering\end{minipage}
\begin{minipage}{0.50\textwidth}
\includegraphics[width=1.9\textwidth]{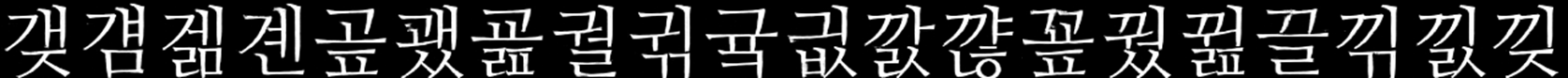}
\end{minipage}\\
\begin{minipage}{0.18\textwidth}\textit{target}\centering\end{minipage}
\begin{minipage}{0.50\textwidth}
\includegraphics[width=1.9\textwidth]{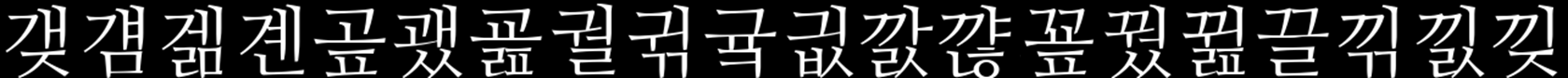}
\end{minipage}\\
}
\end{minipage}
}

\vspace{-7pt}
\ruleh
\hspace{-90pt}
\subfigure{
\begin{minipage}{0.8\textwidth}{
\begin{minipage}{0.18\textwidth}\textit{font2}\centering\end{minipage}
\begin{minipage}{0.50\textwidth}
\includegraphics[width=1.9\textwidth]{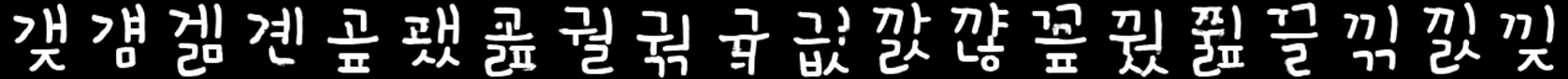}
\end{minipage}\\
\begin{minipage}{0.18\textwidth}\textit{target}\centering\end{minipage}
\begin{minipage}{0.50\textwidth}
\includegraphics[width=1.9\textwidth]{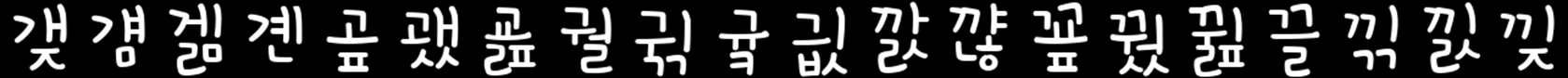}
\end{minipage}\\
}
\end{minipage}
}

\vspace{-7pt}
\ruleh
\hspace{-90pt}
\subfigure{
\begin{minipage}{0.8\textwidth}{
\begin{minipage}{0.18\textwidth}\textit{font3}\centering\end{minipage}
\begin{minipage}{0.50\textwidth}
\includegraphics[width=1.9\textwidth]{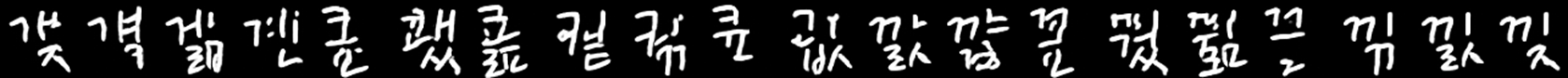}
\end{minipage}\\
\begin{minipage}{0.18\textwidth}\textit{target}\centering\end{minipage}
\begin{minipage}{0.50\textwidth}
\includegraphics[width=1.9\textwidth]{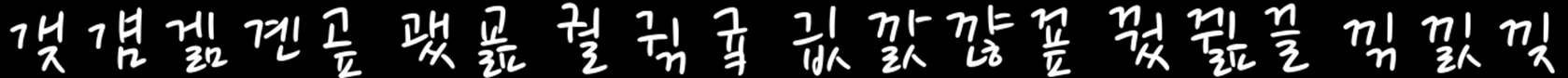}
\end{minipage}\\
}
\end{minipage}
}

\caption{More results of Korean fonts. We randomly select 750 characters for each font to train the model. The 1st row is printed font and the 2nd and 3rd rows are handwritten ones. The results are impressive and they prove that the our model can be used for other hieroglyphics.}
\label{fig: 3}
\end{figure*}

As shown in Figure \ref{fig: 3}, we display results of Korean fonts. The 1st row is printed font and the 2nd and 3rd rows are handwritten ones. We randomly select 750 characters for each font to train the model and performance is pretty good in first two rows. It is because the training samples are selected randomly, the 3rd row is not as good as expected. In Chinese font generation task, we build the small training set $S_D$ by selecting the single-element characters. However, For Korean, we do not use such prior knowledge, so the performance is impacted seriously by the randomness of choice.

\subsection{Appendix \uppercase\expandafter{\romannumeral3}}

\begin{figure*}[ht]
\subsection{Appendix \uppercase\expandafter{\romannumeral3}}
\centering
\includegraphics[width=0.60\textwidth]{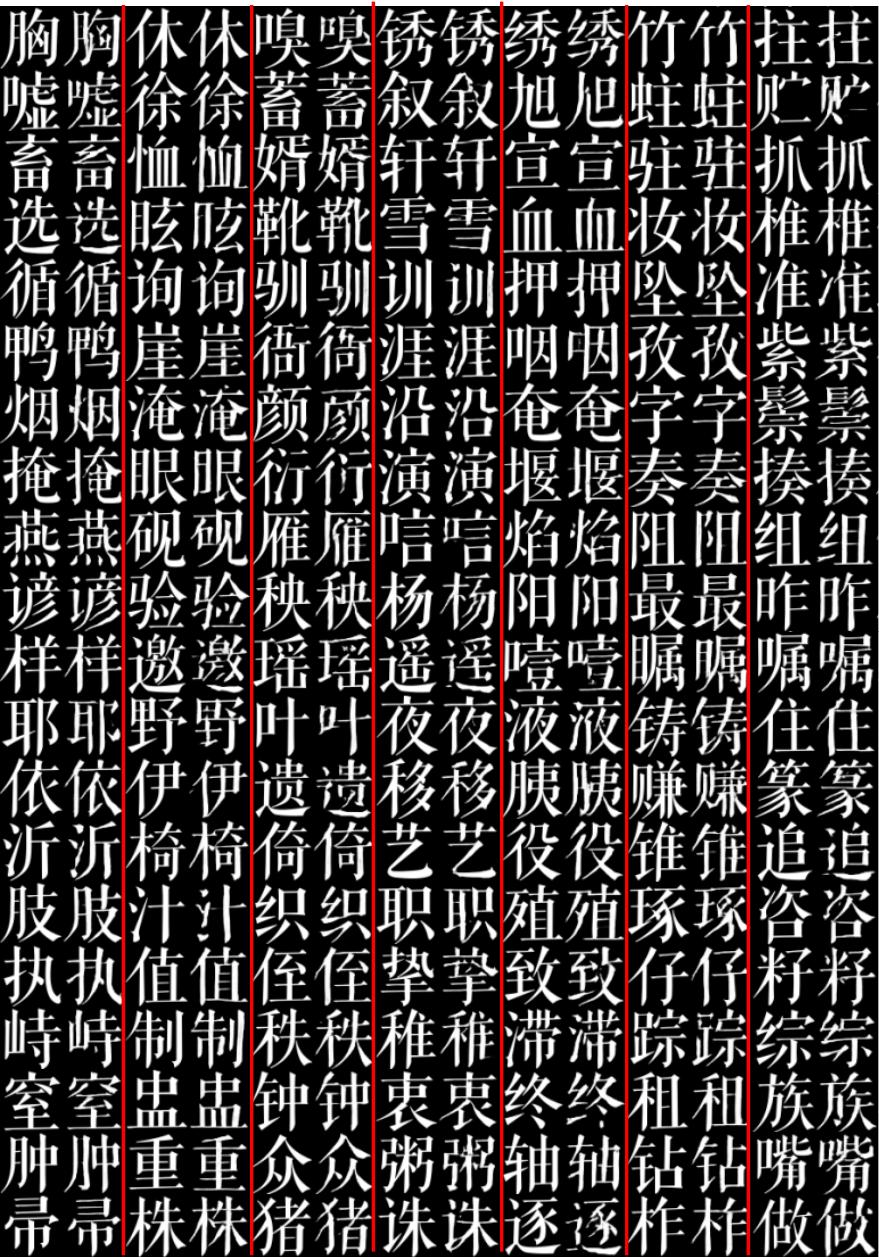}
\caption{The display of a large number of results for Font 1. The characters in the left of each column are ground truths while those on the right are generated characters.}
\label{fig: 4}
\end{figure*}

As shown in Figure \ref{fig: 4}, Font 1 is a printed style and the generated characters are similar to targets except a little blur in details. Our model performs sufficiently well to be used in practical font design.

\begin{figure*}[ht]
\centering
\includegraphics[width=0.60\textwidth]{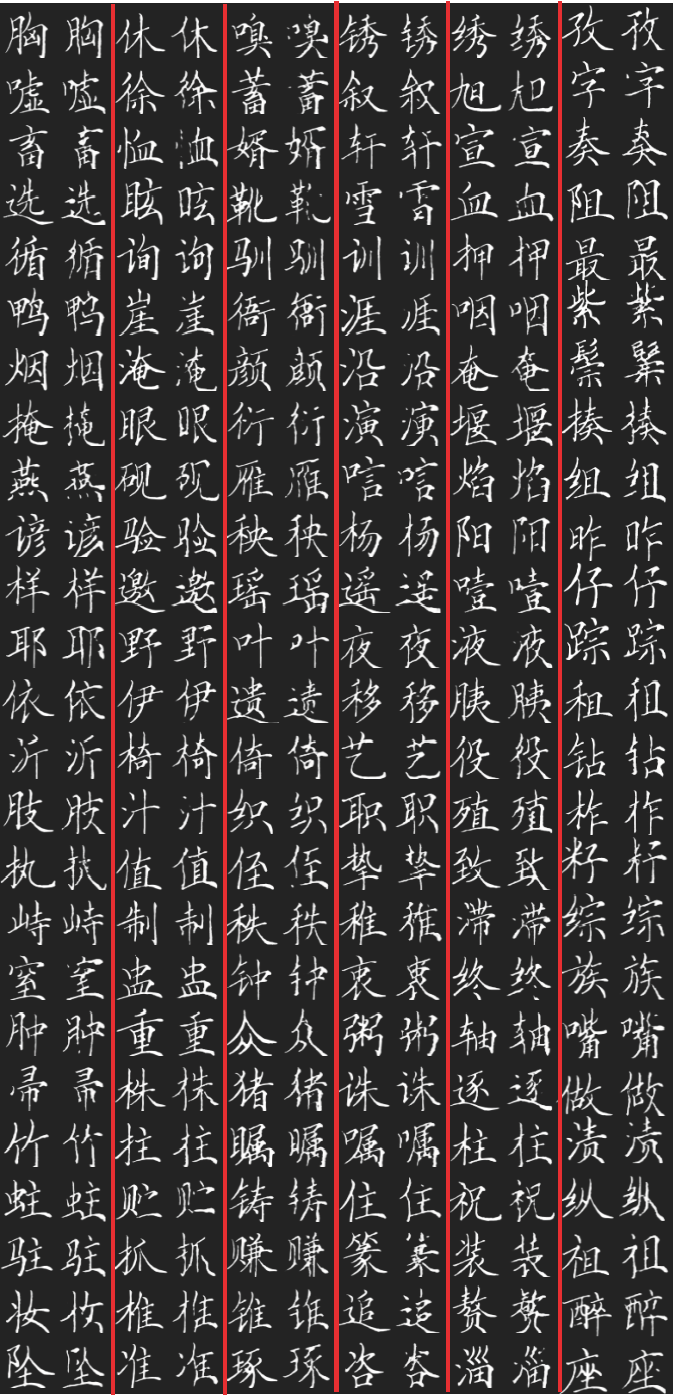}
\caption{The display of a large number of results for Font2. The characters in the left of each column are ground truths while those on the right are generated characters.}
\label{fig: 5}
\end{figure*}

As shown in Figure \ref{fig: 5}, the characteristics of Font 2 is that the strokes are distorted, so the results are not as accurate as Font 1. Even so, the quality of generated characters is still impressive.

\begin{figure*}[ht]
\centering
\includegraphics[width=0.60\textwidth]{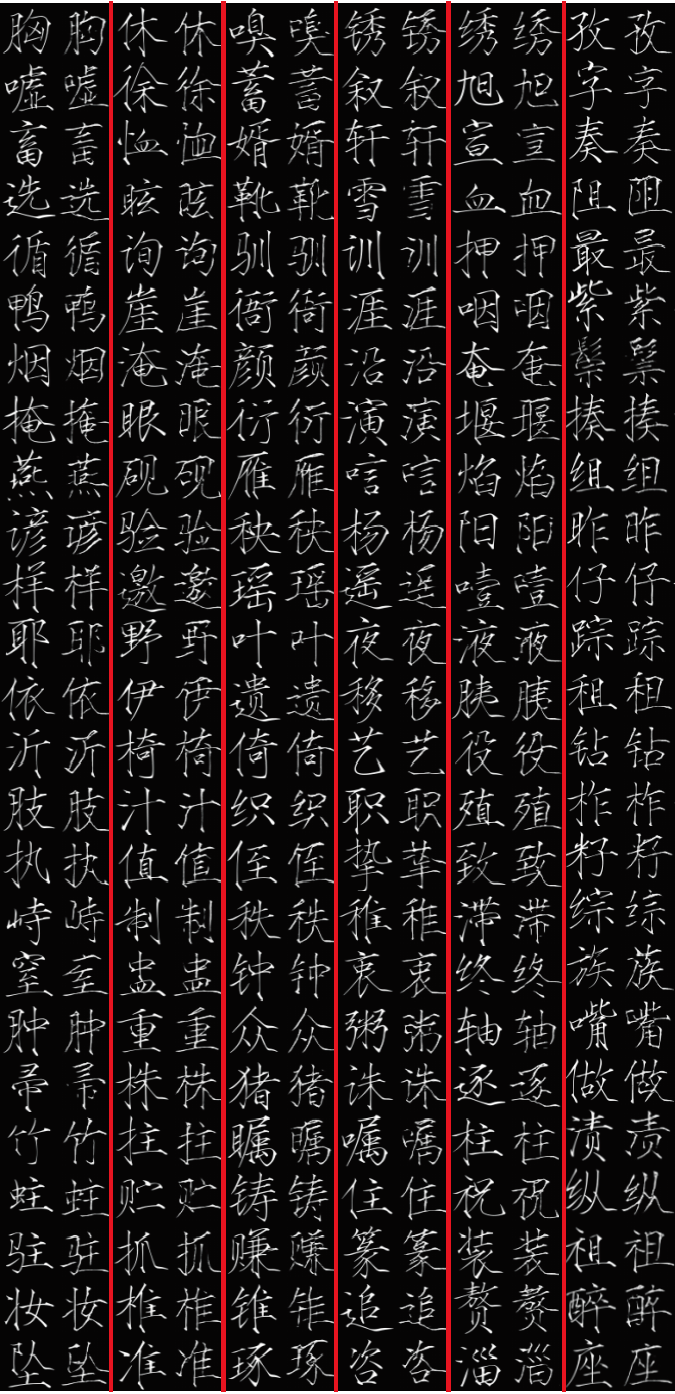}
\caption{The display of a large number of results for Font 3. The characters in the left of each column are ground truths while those on the right are generated characters.}
\label{fig: 6}
\end{figure*}

As shown in Figure \ref{fig: 6}, the strokes of Font 3 is so thin that the outputs of previous methods are usually broken. But our model solves this problem efficiently by Collaborative Stroke Refinement.

\begin{figure*}[ht]
\centering
\includegraphics[width=0.60\textwidth]{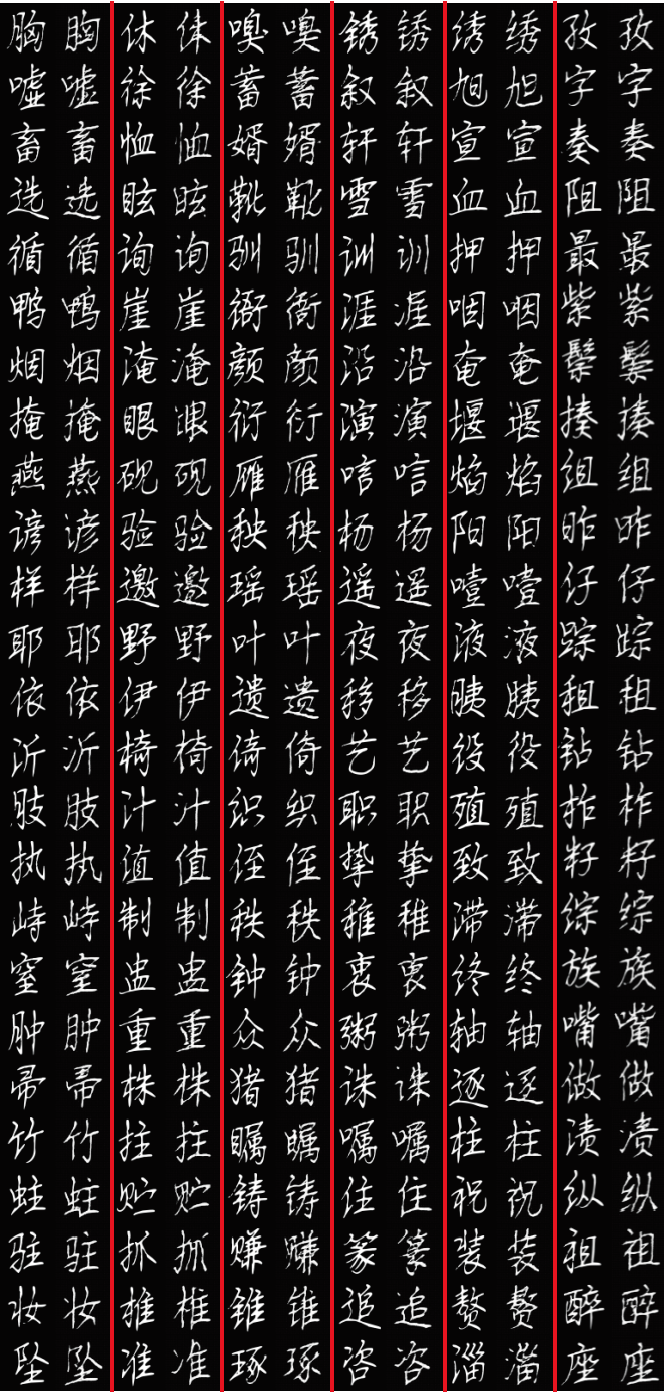}
\caption{The display of a large number of results for Font4. The characters in the left of each column are ground truths while those on the right are generated characters.}
\label{fig: 7}
\end{figure*}

As shown in Figure \ref{fig: 7}, Font 4 is a typical handwritten font. The results are realistic because our model learns the stroke deformation well. The previous models hardly perform well on such irregular fonts.

\subsection{Appendix \uppercase\expandafter{\romannumeral4}}

\begin{figure*}[ht]
\subsection{Appendix \uppercase\expandafter{\romannumeral4}}
\centering
\includegraphics[width=0.85\textwidth]{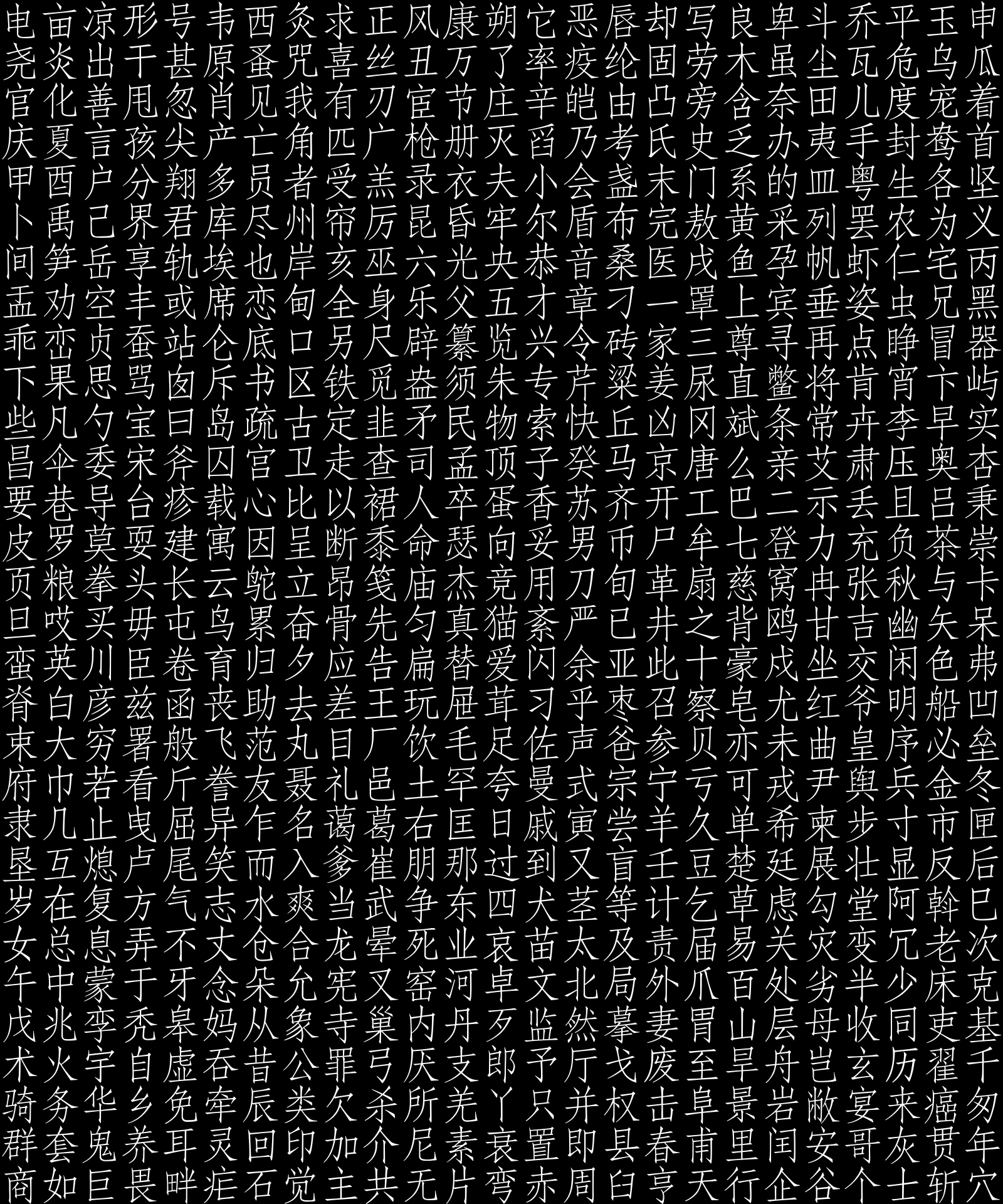}
\caption{The manually selected training set with 750 characters.}
\label{fig: 8}
\end{figure*}

As shown in Figure \ref{fig: 8}, we display the small training set $S_D$ used in all the Chinese font generation experiments in this paper. According to the selection principle,  we select 450 single-element characters and 150$\times$2 compound characters covering all 150 radicals, to create the small dataset $S_D$ totally including 750 training samples. By utilizing the content-reuse phenomenon, we reduce the training set from 3000 to 750, which makes the model practical in font design.

\end{document}